\documentclass{article}
\usepackage{amsmath}
\usepackage{graphicx}
\usepackage{hyperref}

\newcommand{\la} {\langle}
\newcommand{\ra} {\rangle}
\newcommand{\goesto} {\rightarrow}
\title{Testing GPT-4 with Wolfram Alpha \\
and Code Interpreter plug-ins \\
on math and science problems}

\author{Ernest Davis \\ New York University \\ davise@cs.nyu.edu \\
\and
Scott Aaronson\footnote{On leave at OpenAI, 2022-24} 
\\ University of Texas at Austin \\ aaronson@cs.utexas.edu}

\begin{document}
\maketitle

\begin{abstract}
This report describes a test of the large language model GPT-4 with the 
the Wolfram Alpha and the Code Interpreter plug-ins on 105 
original problems in science and math, at the high school and college
levels, carried out in June-August 2023. 
Our tests suggest that the plug-ins significantly enhance GPT's
ability to solve these
problems.  Having said that, there are still often "interface"
failures; that is, GPT often has trouble formulating problems in a way
that
elicits useful answers from the plug-ins.  Fixing these interface
failures seems like a central challenge in making GPT a reliable tool
for college-level calculation problems.
\end{abstract}

This report describes a test of the large language model (LLM)
GPT-4 with the plug-ins
Wolfram Alpha (henceforth ``GPT4+WA'')  and Code Interpreter (henceforth
``GPT4+CI'') on 105 expert-crafted,
original problems in science and math, at the high school and college
levels, carried out in June-August 2023.\footnote{It was later noticed, in September
2024, that problems 3 and 17 of the ``Arbitrary Numerical'' problems
were misworded and that we had miscalculated the answer for
problem 3. The corrections are noted on p. 18 and 24. The correction
of these errors does not affect the evaluation of the AI systems; on both of these
problems, the AI systems made errors much graver than can be attributed to our 
miswordings.}
Table~\ref{tabExamples} shows
examples of the problems used: three where both AI systems succeeded and
three where both failed. The complete collection of problems are in appendices 
A-C; the complete outcomes of the two AIs are in 
tables~\ref{tab1}-\ref{tabMotivatedResults}. 

\begin{table}
\begin{center}
{\bf Successes}  

\vspace{6pt}

Both GPT4+WA and GPT4+CI got the correct answer on these.
\end{center}

\vspace{6pt}

{\bf Problem, ``Arbitrary Numerical'' test set:} A point {\bf p} is chosen at random within the 100-dimensional box
${\bf B}=[0,100]^{100}$ following a uniform distribution. 
What is the probability
that the Euclidean distance from {\bf p} to the boundary of $\bf B$ 
is less than 1? 

{\bf Answer:} $1-0.98^{100} \approx 1-1/e^{2} = 0.8647$

\vspace{6pt}

{\bf Problem, ``Calculation-free'' test set:} Let C be the center of the earth.
Can there be three earth satellites X, Y, and Z such that C, X, Y, and Z
are always coplanar? \\
{\bf Answer:} Yes.

\vspace{6pt}

{\bf Problem, ``Motivated Numerical'' test set:} 
What is the probability that a randomly-chosen 100 $\cdot$ 100 matrix,
over the finite field $F_{2}$, is invertible?
Please include all calculations.

{\bf Answer} 
\[ \prod_{i=1}^{100} \frac{2^{j}-1}{2^{j}} \approx 0.289 \]

\rule{5in}{1pt}

\begin{center}
{\bf Failures}  \\
\end{center}

{\bf Problem, ``Arbitrary Numerical'' test set:} 
Viewed from Vega, what is the angle between Sirius and the Sun?
\\
{\bf Correct Answer:} 0.0972 radians = $5.6^{\circ}$.  {\bf GPT4+WA:} 
.005060 degrees. \\
{\bf GPT4+CI:} ``The angle between Sirius and the Sun as viewed 
from Vega depends on the time of year, as well as the specific time of day. 
This is because the Earth and Vega both orbit the Sun, but
at different rates, and because the Earth also rotates on its axis. \ldots''

\vspace{6pt}

{\bf Problem, ``Calculation-free'' test set:} 
Joe says that he lives 10 miles from Lake Michigan, that 
Beth lives 
10 miles from Lake Michigan,
and that he and Beth live 100 miles apart. Is it possible that 
Joe is telling the truth? Answer `Yes' or `No'.'' \\
{\bf Correct Answer:} Yes. {\bf GPT4+WA and GPT4+CI both answer:} No.

\vspace{6pt}

{\bf Problem, ``Motivated Numerical'' test set:} 
What is the Shannon entropy of a positive integer $n$ that's chosen
with probability Pr[n] = $6/(\pi^2 \cdot n^2)$?
Please include all calculations. \\
{\bf Answer:} 2.362. {\bf GPT4+WA:} did not return an answer.
{\bf GPT4+CI:} 0.9265.
\caption{Examples of successes and failures on test set problems}
\label{tabExamples}
\end{table}

\section{Summary of Conclusions}

Our test sets were too small and too haphazard to support statistically valid
conclusions, but they were suggestive of a number of conclusions. We
summarize these here, and discuss them at greater length 
in section~\ref{secDiscussion}.

Over the kinds of problems tested, GPT-4 with either plug-in is significantly
stronger than GPT-4 by itself, or, almost certainly, than any AI that existed
a year ago.  However it is still far from reliable; it often outputs a wrong
answer or fails to output any answer.

In terms of overall score, we would judge that these systems performs on the 
level of a middling undergraduate student. However, their capacities and
weaknesses do not align with a human student; the systems solve some 
problems that even capable students would
find challenging, whereas they fail on some problems that even middling
high school students would find easy.


There is considerable room for improvement with the interfaces between GPT-4 
and the plug-ins, particularly with Wolfram Alpha. GPT-4 often struggles to
formulate a problem in a way that Wolfram Alpha can accept or that produces
useful output. GPT-4 also sometimes creates quite useless calls to
the plug-ins.

GPT-4 fails to take full advantage of the capacities of the plug-ins, 
particularly Wolfram Alpha.
The LLM often carries out calculations, retrieves formulas, and does symbolic
manipulations that the specialized plug-ins could carry out more reliably.
This sometimes leads to errors that should have been avoidable.

Overall, the systems are strongest on problems that can be solved by invoking
a single formula. They are often weak on problems that humans would tend to 
solve using spatial visualization. They are often weak on problems that involve
combining several calculations of different kinds.  The systems often have
trouble dealing with numbers that are very large or very small.

GPT-4 has some ability to detect when the answer returned by a plug-in is
nonsensical or physically meaningless, but not reliably, and it has little
ability to diagnose the reason for the mistake or to recover from it.

\section{The test sets: Overview}
\label{secTestSets}
We created three test sets of original problems of somewhat different flavors
which we here call the ``Arbitrary Numerical'' test set, the ``Calculation-Free''
test set, and the ``Motivated Numerical'' test set.  The problems in the first
two were written by Davis, the third was written by Aaronson; they
reflect the interests and tastes of their authors.
The complete collection
of problems with answers are in appendices A, B, and C below.\footnote{The 
problems, answers, and the output from the two AIs, edited and annotated, 
are on the web:
\url{https://cs.nyu.edu/~davise/papers/GPTPlugInTests/}}

\subsection{The ``Arbitrary Numerical'' test set: Overview}
The first test set, which for convenience we will call the ``Arbitrary 
Numerical'' 
test set, contains 32 problems, whose answers are either numerical or,
in four cases, two- or three-dimensional vectors. Other than numerical 
constants, which are easily available in standard references such as Wikipedia,
and are known by both GPT-4 and Wolfram Alpha, the science involved in
this dataset is quite elementary. There are two questions that require using
the relativistic formula for energy and two that involve Planck's law relating
the wavelength and energy of a photon; otherwise, the science required is
all at the level of high school or first-semester college physics and 
chemistry. The mathematics demanded is somewhat more demanding, drawing
on such subjects as three-dimensional geometry, spherical geometry, linear 
algebra, probability theory, integral calculus, and, in one problem, asymptotic
analysis; but only very elementary aspects of these fields.
The problems are ``arbitrary'' in the sense that each problem involves a
fairly random collection of specifics and that the individual problems are
not of any particular inherent interest.
Some of the problems would be considered somewhat complex and, frankly, 
rather tedious for a human student, but none involve any very deep reasoning.

\subsection{The ``Calculation-Free'' test set: Overview}
The second test set, called 
the ``Calculation-Free'' test set, contains 53 questions with
discrete answers: 36 problems with two possible answers, 10 multiple choice 
problems, and 7 sorting problems. Since a reasonable score can be gotten 
by guessing, we included multiple instances of problems of the same general
type. In general a human test-taker who has a general knowledge of the
subject matter and who either knows or can look up the geographical and
historical information involved can solve these without using a calculator
or setting pencil to paper.

The first category, ``Eclipses'' consists of eight 
multiple-choice questions about what an astronomer on the 
moon would observe during an eclipse. For example: 
\begin{quote}
An astronaut is standing in the Sea of Tranquility during what on earth is called a total lunar
eclipse. They are looking in the direction of the earth. What they see is: \\
A. The surface of the moon, illuminated by earth light. \\
B. The night side of the earth, occluding the sun. \\
C. The surface of the moon, illuminated only by starlight. \\
D. The surface of the moon, illuminated by the sun. \\
E. The sun. \\
F. The day side of the earth, with a small circular shadow moving quickly over it. \\
G. The night side of the earth. The sun is somewhere else entirely. \\
H. A starry sky. Neither the sun, the earth, or the surface of the moon is in the field of view. \\
\end{quote}
(The answer is B.)
Anyone with an elementary-school understanding of the relation of the earth,
moon, and sun during eclipses should be able to answer these without 
difficulty.

The second category, ``Distance combinations''
consists of seven questions along the following lines,
``Joe says that he lives 10 miles from Lake Michigan, that Beth lives 
10 miles from Lake Michigan,
and that he and Beth live 100 miles apart. Is it possible that 
Joe is telling the truth? Answer `Yes' or `No'.''

Anyone who knows, or can look up on a map, the geography of the
Atlantic Ocean, Lake Michigan, Lake Huron, and Walden Pond (the four bodies of
water mentioned) can easily solve these.

The third category, ``Distance between points in rivers'', contains five 
questions 
such as the following: ``Is there a point x in the Danube River and a point y in the Rhine River that are exactly 429 miles apart? Answer `Yes' or `No'.''

Anyone who is familiar with the geography of the Nile, Danube, and Rhine
rivers, or can look them up on a map, can answer these easily.

The fourth category, ``Clockwise or counter-clockwise'' presents a triple
of Western Hemisphere cities, such as ``Caracas, Venezuela; 
Amarillo, Texas; Quebec, Quebec,''
 and asks if the sequence is in clockwise or counter-clockwise order. 
Anyone who can visualize the geography involved or can look up a map can
answer these easily. There are twelve of these questions.

The fifth category, ``Sorting'' contains seven problem that ask 
the users to sort specified objects or events
(between 8 and 11) in increasing order by mass, temporal characteristics,
or distance from a reference point. For example, 
\begin{quote}
Sort the events below by starting date: \\
A. The lifetime of Marie Antoinette. \\
B. The Precambrian period. \\
C. The first performance of Beethoven’s seventh symphony. \\
D. Lincoln speaking the Gettysburg address. \\
E. The Hundred Years’ War. \\
F. The reign of Queen Victoria. \\
G. The existence of the species of passenger pigeons (ending with the death of “Martha”). \\
H. The battle of Gettysburg. \\
I. The existence of legal slavery in what is now the United States. \\
J. The reign of Marie Antoinette. \\
K. The lifetime of Joan of Arc. 
\end{quote}

Due to a misdesign, three of specific comparisons in the seven problems turned
out to be tricky, 
but the rest could be easily judged by someone who knows
or can look up the relevant history and geography.

The sixth category ``Satellites''
consists of fourteen problems about the feasibility of 
Earth satellite orbits with various characteristics. Twelve of these are binary
choice; two are multiple-choice. For example, ``Is it possible to have 
a satellite such that the northernmost earth point underneath the 
satellite has latitude 40 degrees N and the southernmost has latitude 
30 degrees S? Assume that the satellite is moving in a closed orbit 
around the Earth and that the only influence on the satellite's motion  
is the Earth's gravity. Assume that the Earth is a perfect sphere. 
Ignore the revolution of the Earth around the sun, but do not ignore 
the rotation of the Earth around its axis.''

These require an understanding of the basic properties of satellite orbits and 
some ability to reason about three-dimensional geometry. For example, answering
the question above requires understanding that the plane of a satellite's orbit
always includes the center of the earth, and that that implies that the
the southernmost point will have equal latitudes with opposite signs. However,
no calculations are ever involved.

\subsection{The ``Motivated Numerical'' test set: Overview}
The third test set, called the ``Motivated Numerical'' test set, consists
of 20 problems with numerical answers. These were designed to be more 
natural and of greater inherent interest than the ``Arbitrary Numerical''
problems; one can imagine someone raising one of them in casual geeky 
conversation as an interesting factoid, which is rarely the case for the
arbitrary problems. They draw on a wide range of areas in math and physics, 
often somewhat more advanced than the arbitrary problems.

\section{History of the AI systems and the testing project}
\label{secHistory}
AI technology currently advances (and occasionally retreats)
so rapidly that it has become necessary to
report precise dates for experiments, and that even quite short and small 
projects can be upended in midstream by new external developments.

On March 14, 2023, OpenAI (2023a, 2023b) released the LLM GPT-4.
On March 23, OpenAI (2023c)
that GPT-4 would support plug-ins. The same day Steven Wolfram (2023)
announced the availability of a plug-in that would allow GPT-4 to 
interface with the ``answer engine'' Wolfram 
Alpha.\footnote{\href{http://wolframalpha.com} {Wolfram Alpha}
supports very powerful algorithms for mathematical computation, particularly
in symbolic math, and has extensive databases of scientific, engineering,
and other encyclopedic data. It was released in May 2009. It accepts input
in both symbolic form and in natural language (English); the natural 
language interface, however, is quite limited.} The announcement gives 
considerable detail and many examples. Plug-ins are currently
considered a beta-level release; they are
available to paying subscribers to ChatGPT-Plus.

It seemed to us likely that the combination of GPT-4 with Wolfram Alpha 
might be significantly more powerful for science and math problems posed
in English than earlier systems; it certainly seemed worth testing. 
We developed three test sets of original problems, described in 
section~\ref{secTestSets} below, and included in full in 
appendices A, B, and C. 
We began testing GPT4+WA  on June 28. By July 9 we had nearly completed the
testing and analysis of the first two test sets.

On July 9 we learned that, three days earlier, on July 6, OpenAI had 
announced
the beta-level release of
a plug-in ``Code Interpreter'' which enables GPT-4 to write and execute
code in Python, with access to Python's extensive mathematical libraries. 
It was clear that our
tests should also be run on the GPT4+CI;
we carried out those tests of GPT+CI
on the first two data sets between July 9 and
July 12.  The third data set was tested between July 27 and August 2
(table~\ref{tabDates}).
It is these tests that are described in this report.

\begin{table}
\begin{center}
\begin{tabular}{|l|c|c|c|} \hline
        & {\bf "Arbitrary Numerical"} & {\bf ``Calculation-Free''} &
{\bf ``Motivated Numerical''} \\ \hline
{\bf GPT4+WA} & 6/28-7/9 & 7/9-7/10 & 7/27-8/2 \\ \hline
{\bf GPT4+CI} & 7/10-7/12 & 7/10-7/12 & 7/27-8/2\\ \hline 
\end{tabular}
\end{center}
\caption{Dates of experiments}
\label{tabDates}
\end{table}

\section{The Design and Testing Process}
The three sets of questions, and their answers, were all prepared in advance
of any testing; the authors had no experience using GPT-4 with the plug-ins.
Some edits for disambiguation were found to be 
necessary in the
course of testing, and consequently some problems were run more than once,
with different wordings. For example: In problem 3 of the Arbitrary Numerical 
test set, we initially omitted the unit length of the tetrahedron. 
Problem 19 of the Arbitrary Numerical test set was originally worded ``\ldots
vertex B is
in the x-y plane''; we changed that to ``in the x-z plane, with positive
x-coordinate'' since otherwise the problem is ambiguous. In the ``Satellite''
problems of the ``Calculation Free''
test set, we added a specification of the idealizations
to be applied. However, all of these were driven by considerations of
the problems themselves, not in reaction to the results from our tests; and
no contentive changes were made.

One problem was deleted, after running it through the AI systems, because
we discovered that the correct answer is an open problem in coding 
theory.\footnote{Specifically, the problem was ``What is, roughly, the maximum 
number of 100-bit strings that can be
chosen, such that every pair of them has Hamming distance at least 40?''.
This problem is still one of active research; see (Ye et al. 2023) for
recent results. A well-known lower bound for this problem is the
Gllbert-Varshamov bound, which for the values $n=100, d=40$ has a value
of 56. However, this is by no means a tight bound; it is not difficult
to show that the true value is at least 128. Both GPT4+WA and GPT4+CI
returned the Gilbert-Varshamov bound, gave a correct explanation, and
evaluated it correct; however, GPT4+CI presented it as the exact
answer to the question, and GPT4+WA claimed that the true value might be
slightly larger.}

To avoid cross-question contamination, we started a new Chat session for
each of the problems. The only exception was the ``clockwise'' problems,
where that seemed burdensome and unnecessarily; 
rather, for the first question we wrote,
\begin{quote}
If you have a map that shows Chicago, New York City, and Atlanta, and 
you draw a circle through the three of them, then the sequence 
$<$ Chicago; New York; Atlanta $>$ is in clockwise order. The sequence $<$
New York; Chicago; Atlanta$>$ on the other hand, is in counterclockwise order.

Is the following in counterclockwise or counterclockwise order:  Caracas, Venezuela; Amarillo, Texas; Quebec, Quebec.
\end{quote}
and then for the subsequent questions we continued the session and asked,
\begin{quote}
How about New Orleans, Louisiana; Springfield, Illinois; Jacksonville, Florida.
\end{quote}
and so on.

Obviously, this collection of problems is quite unsystematic and arbitrary,
as well as small, and it does not support any broad conclusions about the 
capabilities, absolute or relative, of these two systems. Even if one
system had achieved a perfect success on these test sets, and the other
had failed abjectly, that would hardly justify concluding that the first
is better in general than the second, let alone that the first ``understands''
the subject matter and the second does not.  However, the
results are suggestive of some interesting strengths and weaknesses.

We did not experiment with few-shot prompts or ``chain of thought'' prompts.
(In most cases, the two systems output a chain of thought in any case.)
In many of our tests of the Arbitrary Numerical test set we ended our prompt.
``Please include
all calculations in LaTeX, suitable for copying and pasting;'' the system
sometimes complied with that format request, but not at all reliably.
In all of our tests of the ``Motivated Numerical'' test set, we ended our
prompt with ``Please include all calculations''.  

\section{Results}
\label{secResults}

Tables~\ref{tab1}-\ref{tabMotivatedResults} 
give a tabular summary of the results of our
tests.

\begin{table}
We use the error marking ``Wrong analysis'' in cases where, if a human being
had produced the output, a grader would be inclined to say that they had
understood the problem conceptually but had erred in setting up the mathematical
analysis. We use the error marking ``Misunderstood problem'' in cases where a
grader would say that a human really did not understand, either what was the
specification or what was being asked. Obviously, the borderline between these
is indeterminate; also obviously, applying them to AIs is reckless 
anthropomorphism, meant only to be suggestive.

In the categorization of the science and math required, we focus on the most
important component of the analysis. 

\vspace{6pt}

\begin{tabular}{|l|l|l|l|l|l|} \hline
\#  & Science & GPT4+  & GPT4+WA error & GPT4+  & GPT4+CI error \\ 
    & and math &WA score   &               &CI score    &               \\ \hline
1   & Gravity  & 1 &                     &  1 &                     \\ 
    & 3D geometry &  &                   &    &                    \\ \hline 
2   & 3D geometry & 0 & Misunderstood problem & 0 & Error in formula \\ \hline
3   & Special relativity & 0 & Used non-relativistic & 0 & Used non-relativistic 
\\ 
    & Asymptotics &  & formula               &   & formula    \\  \hline
4.  & Stellar data       & 0 & Misunderstood & 0 & No answer \\
    & 3D geometry        &   & problem        &   & Misunderstood  \\ \hline
5.  & Basic physics      & 0 & Wrong analysis & 1 &               \\ 
    & 3D geometry        &   &                &   &               \\ \hline
6.  & Gravity            & 0 & Unclear       & 1 &                \\
    & Vectors            &   &               &   &                \\ \hline
7.  & Table of eclipses  & 0 & Completely    & 0 &  Didn't answer \\
    & Counting           &   & lost          &   &                \\ \hline
8.  & Ideal gas law      & 1 &               & 1 &                \\
    & 3D geometry        &   &               &   &                \\ \hline
9.  & Doppler effect     & 0 & Misunderstood & 0 & Erroneous      \\ 
    & Algebra            &   & problem       &   & formula        \\ \hline
10. & Planck's law       & 0.75 &             & 0 & Didn't answer  \\
    & Divergent integral &   &               &   &                \\ \hline
11. & Planck's law       & 0 & Wrong analysis & 1 &               \\ 
    & Probability        &   &                &   &               \\ \hline
12. & Mechanics          & 0 & Misunderstood & 0 & Misunderstood  \\
    & Algebra + trig     &   & problem        & 0 & problem        \\ \hline
13. & Solar system       & 0 & Wrong analysis & 1 &               \\ 
    & 3D geometry        &   &                &  &                \\ \hline
14. & Gravity            & 0.75 & No answer      & 0 & Error in      \\
    & Algebra            &   & Unable to      &   & formula       \\
    &                    &   & convert units  &   &               \\ \hline
15. & Spherical          & 0 & Misunderstood  & 0 & Gave trivial  \\
    & geometry          &   & problem        &   & upper bound   \\ \hline
16. & Isotope data       & 0 & Wrong formula  & 0 & Miscalculated? \\ 
    & Probability        &   &                &   &                \\ \hline
\end{tabular}
\caption{Results for scientific problems: 1-16}
\label{tab1}
\end{table}

\begin{table}
\begin{tabular}{|l|l|l|l|l|l|} \hline
\#  & Science & GPT4+  & GPT4+WA error & GPT4+  & GPT4+CI error \\ 
    & and math & WA score   &          & CI  score    &               \\ \hline
17. & Coulomb's law & 0  & Sign error    & 0        & Sign error    \\
    & Vectors       &    & Physics error &          & Vector error  \\  \hline
18. & Trig          & 0 &  Wrong analysis & 0       & Wrong analysis \\ \hline
19. & 3D geometry   & 0  & Wrong analysis & 0       & Wrong analysis  \\
    & Linear algebra &   &                &         &                \\ \hline
20. & Compound      & 1  &                & 0 & Didn't call CI  \\
    & interest      &    &                &   & Wrong analysis \\ \hline
21. & Probability   & 0.75 & Right answer & 0 & Used Monte Carlo \\
    &               &    & but WA couldn't & & for tiny probability \\
    &               &    & simplify        & &                      \\ \hline
22. & Probability   & 1  &                & 1 &                     \\ \hline
23. & Probability   & 0  & Wrong analysis & 0 & Wrong analysis      \\ 
    & Calculus      &    &                &   &                     \\ \hline
24. & Chemistry     & 0  & Misunderstood  & 0 & Wrong analysis      \\ 
    & Algebra       &    & problem        &   &                     \\ \hline
25. & Radioactivity & 1 &  Mistake but    & 1 &                     \\
    & Algebra       &   & better than     &   &                     \\ \hline
    &               &   & author          &   &                     \\ \hline
26. & Probability   & 0 & Wrong formula   & 1 &                     \\ \hline
27. & Probability   & 0 & Misunderstood   & 0 & Misunderstood       \\
    &               &   & problem         &   & problem             \\ \hline 
28. & Probability   & 0 & Unintelligible  & 0 & Unintelligible      \\
    &               &   & calculation     &   & calculation         \\ \hline 
29. & Satellite     & 0 & Wrong analysis  & 0 & Wrong analysis      \\ 
    & 3D geometry   &   &                 &   &                     \\ \hline
30. & Satellite     & 0 & Wrong analysis  & 0 & Wrong analysis      \\ 
    & 3D geometry   &   &                 &   &                     \\ \hline
31. & Satellite     & 1 &                 & 1 &                     \\ 
    & 3D geometry   &   &                 &   &                     \\ \hline
32. & Satellite     & 0 & Misunderstood   & 0 & Misunderstood       \\
    & 3D geometry   &   & problem         &   & problem             \\ \hline
\end{tabular}

In total, GPT4+WA got a score of 8.25 out of 32, giving partial credit on
three problems; GPT4+CI got a score of 10 out of 32. 
\caption{Results for scientific problems: 17-32}
\label{tab2}
\end{table}

\begin{table}
\begin{tabular}{|l|l|l|l|l|l|l|} \hline
\# & \multicolumn{3}{c|}{GPT4+WA} & \multicolumn{3}{c|}{GPT4+CI} \\ 
   & Score  & Explanation & called WA  & Score & Explanation  & called CI \\ 
\hline
\multicolumn{7}{|c|}{Eclipses. Expected value guessing randomly: 1/8} \\ \hline
1.  & 0 & Wrong    &    & 0 & Wrong      &       \\ \hline
2.  & 1 & Right    &    & 1 & Right      &       \\ \hline
3.  & 0 & Wrong    &    & 1 & Right      &       \\ \hline
4.  & 0 & Wrong    &    & 0 & Wrong      &       \\ \hline
5.  & 0 & Wrong    &    & 0 & Wrong      &       \\ \hline
6.  & 0 & Wrong    &    & 0 & Wrong      &       \\ \hline
7.  & 0 & Wrong    &    & 0 & Wrong      &       \\ \hline
8.  & 1 & Right    &    & 1 & Right      &        \\ \hline

\multicolumn{7}{|c|}{Distance combination problems. Random guessing: 3.5/7} \\
\hline
9.  & 1 & Right    &    & 0 & Gibberish  &       \\ \hline
10. & 1 & Small flaw &  & 1 & Small flaw &       \\ \hline
11. & 0 & Wrong      &  & 0 & Wrong      &       \\ \hline
12. & 1 & Right      &  & 1 & Right      &       \\ \hline
13. & 0 & Wrong      &  & 0 & Wrong      & Yes  \\ \hline
14. & 1 & Wrong      &  & 1 & Wrong      &      \\ \hline
15. & 0 & Wrong      &  & 0 & Wrong      &      \\ \hline
\multicolumn{7}{|c|}{Distance between points in rivers. Random guessing: 2.5/5}
\\ \hline
16. & 1 & OK     & Yes  & 0 & None       &      \\ \hline
17. & 1 & Right  & Yes  & 1 & None       &      \\ \hline
18. & 1 & Right  & Yes  & 1 & None       &      \\ \hline
19. & 1 & Right  & Yes  & 1 & Right      &      \\ \hline
20. & 1 & Right  & Yes  & 1 & None       &      \\ \hline
\multicolumn{7}{|c|}{Clockwise or counterclockwise. Random guessing: 6/12}
\\ \hline
21. & 0 & None   & Yes  & 1 & None       &      \\ \hline
22. & 1 & None   & Yes  & 0 & None       &      \\ \hline
23. & 0 & None   & Yes  & 1 & None       &      \\ \hline
24. & 1 & None   & Yes  & 0 & None       &      \\ \hline
25. & 0 & None   & Yes  & 1 & None       &      \\ \hline
26. & 0 & None   & Yes  & 1 & None       &      \\ \hline
27. & 0 & None   & Yes  & 1 & None       &      \\ \hline
28. & 1 & None   & Yes  & 1 & None       &      \\ \hline
29. & 0 & None   & Yes  & 1 & None       &      \\ \hline
30. & 1 & None   & Yes  & 1 & None       &      \\ \hline
31. & 0 & None   & Yes  & 1 & None       &      \\ \hline
32. & 1 & None   & Yes  & 0 & None       &      \\ \hline
\end{tabular}
\caption{Results for calculation-free problems 1-32}
\label{tab3}
\end{table}

\begin{table}
Sorting problems are scored in terms of the fraction of correct pairwise 
comparisons. In this category, GPT4+WA always called WA to find the quantities,
with disastrous consequences in questions 35 and 37. Our evaluation of the 
``explanation'' here reflects the accuracy of the AI's estimation of the
quantities involved, rather than the success in sorting those. (The sorting
was always carried out by GPT rather than the plug-in.)

\vspace{6pt}

\begin{tabular}{|l|l|l|l|l|l|l|} \hline
\# & \multicolumn{3}{c|}{GPT4+WA} & \multicolumn{3}{c|}{GPT4+CI} \\ 
   & Score  & Explanation & called WA  & Score & Explanation  & called CI \\ 
\hline
\multicolumn{7}{|c|}{Sorting. Expected value guessing randomly: 3.5/7} \\ \hline
33. & 0.96   & Right     & Yes & 1 & Right     &    \\ \hline
34. & 1      & Small flaw & Yes & 0.98 & Right &   \\ \hline
35. & 0      & No answer  & Yes & 0.96 & Right &   \\ \hline
36. & 0.85   & Right      &     & 0.25 & Right &   \\ \hline
37. & 0      & No answer  & Yes & 0.98 & Small flaw & \\ \hline
38. & 1      & Right      & Yes & 1    & Right      & Yes \\ \hline
39. & 0.857  & Right      & Yes & 1    & Right      & Yes \\ \hline
\multicolumn{7}{|c|}{Satellites. Expected value guessing randomly: 6.08/13} 
\\ \hline
40. & 1      & Right      &     & 0    & Wrong      &      \\ \hline
41. & 1      & Right      &     & 1    & Nonsense   &      \\ \hline
42. & 0      & Nonesense  &     & 1    & Right      &      \\ \hline
43. & 0      & Wrong      &     & 0    & Wrong      &      \\ \hline
44. & 1      & Right      &     & 1    & Right      &      \\ \hline
45. & 1      & OK         &     & 1    & Right      &      \\ \hline
46. & 1      & Right      &     & 1    & Right      &      \\ \hline
47. & 0      & Wrong      &     & 0    & Wrong      &      \\ \hline
48. & 1      & Right      &     & 1    & Right      &      \\ \hline
49. & 0      & Wrong      &     & 0    & Wrong      &      \\ \hline
50. & 1      & Wrong      &     & 1    & Wrong      &      \\ \hline
51. & 1      & So-so      &     & 0    & Wrong      &      \\ \hline
52. & 1      & Right      &     & 1    & Right      &      \\ \hline
53. & 1      & Wrong      &     & 1    & Wrong      &      \\ \hline
\end{tabular}
 
GPT4+WA: 30.7/53.  GPT4+CI: 34.2/53.  Guessing randomly: 22.6/53
\caption{Results for calculation-free problems 35-53}
\label{tab4}
\end{table}

\begin{table}
\begin{tabular}{|l|l|l|l|l|l|} \hline
\#  & Science & GPT4+  & GPT4+WA error & GPT4+  & GPT4+CI error \\ 
    & and math &WA score   &               &CI score    &             \\ \hline
1.  & Black holes & 1      &               & 1          &             \\ \hline
2.  & Black holes & 1      &               & 1          &             \\ \hline
3.  & Black holes & 1      &               & 1          &             \\ \hline
4.  & Stirling's  & 1      &               & 1          &             \\ 
    & formula     &        &               &            &             \\ \hline
5.  & Geometry    & 0      & Irrelevant    & 0.5 & Small angle        \\ 
    &             &        & formula       &     & approximation      \\ \hline
6.  & Geometry    & 1      &               & 0   & Wrong formula      \\ \hline
7.  & Chronology  & 0.5    & Miscalculated & 0.5 & Misinterpreted     \\
    &             &        &               &     & question           \\ \hline
8.  & Information & 1      &               & 1   &                    \\ \hline
9.  & Gravity     & 1      &               & 0   & Symbolic           \\
    &             &        &               &     & manipulation       \\ \hline
10. & Calculus    & 1      &               & 1   &                    \\ \hline
11. & Special     & 0      & Meaningless   & 0.5 & Omitted one        \\ 
    & relativity  &        &               &     & consideration      \\ \hline
12. & Cosmic rays & 1      &               &  1  &                    \\ \hline
13. & Number      & 0      & Simplify arith-   & 1   &                \\ 
    & theory      &        & metic expression &   &                   \\ \hline  
14. & Linear      & 1      &               & 1   &                   \\ 
    & algebra     &        &               &     &                    \\ \hline
15. & Historical  & 1      &               & 1   &                    \\ 
    & estimate    &        &               &     &                    \\ \hline
16. & Information & 0      & No answer     & 0   & Sign error         \\ 
    & theory      &        &               &     &                    \\ \hline
17. & Calculus    & 0      & Wrong formula & 0.5 & Failed to exclude  \\
    &             &        &               &     & incorrect root     \\ \hline
18. & Probability & 0.8    & Ignored 50-50 & 0.8 & Ignored 50-50      \\
    &             &        & split         &     & split              \\ \hline
19. & Combinatorics & 1    &               & 1   &                    \\ \hline
20. & Sphere      & 1      &               & 0   & Geometric          \\
    & packing     &        &               &     & analysis           \\ \hline
\end{tabular}`
\caption{Results from the ``Motivated Numerical'' test set. GPT4+WA: 14.3/20.
GPT4+CI: 13.8/20}
\label{tabMotivatedResults}
\end{table}

\clearpage

\section{Related Work}
\subsection{SciBench}
A similar study has been published very recently by Wang et al. (July 2023).
They created a new benchmark collection ``SciBench'' consisting of
695 problems collected from ten existing college-level textbooks 
in physics,
chemistry, and math, 112 of which had detailed step-by-step
solutions and 104 problems from midterms and exams from courses in data mining,
machine learning, and differential equations. All of the textbook problems,
and most though not all of the exam problems, have numeric answers. 

The SciBench benchmark was used to test GPT-3.5 and GPT-4 in seven
different prompting protocols: Zero-shot with (1) and without (2) the system
prompt; (3) few shot; (4) chain of thought; (5) few-shot with chain of 
thought; 
(6) generating executable Python code using few-shot prompting
(7) generating executable Wolfram Alpha code using few-shot prompting.
The strengths and weakness thus revealed were analyzed by a combination
of human annotators and automated analysis using GPT-3.5.

Obviously, this was a much larger undertaking than our experiment; their test
set is eight times larger than ours, and they ran their experiment on fourteen 
differenti models, while we ran ours on two.
Obviously,
there is some overlap but there are also significant differences in the 
kinds of 
problems in our data sets vs. theirs. In particular, SciBench seems to contains
few problems similar to those in our ``Calculation-Free'' collection and
has a smaller proportion of geometry problems and astronomy problems.
One difference that should
be particularly noted is that, though, like us, Wang et al. tested combinations
of GPT with Python and with Wolfram Alpha, they used their own system of
prompts rather than the plug-ins that have been created by OpenAI and 
Wolfram Alpha. That makes a substantial difference. For instance, Wang
et al. report that ``[Prompts] that utilize Wolfram diminish performance,''
whereas we found that using the Wolfram Alpha plug-in certainly improved
performance over zero-shot GPT-4.

\subsection{Other work}
Many benchmark collections of mathematical word problems have been
assembled  The mathematical difficulty ranges from
elementary school to International Mathematical Olympiad and introductory
college math courses. The language use ranges from non-linguistic to
moderately complex. Some require no world knowledge; others requires some
combination of commonsense, common, expert, and encyclopedic knowledge.
Question formats include open form question answering,
fill-in-the-blank, multiple choice, and proof construction. Some notable
recent examples: SVAMP  (Simple Variations on Arithmetic 
Math Problems) (Patel, Bhattamishra, and Gohal, 2021) 
is a collection of word problems involving
only one or two arithmetic operations and integers between 1 and 999. 
{\sc L\=ila} (Mishra et al. 2022) is a wide-ranging collection, assembled by 
combining twenty-three earlier problem collections; GSM8K (Cobbe et al.
2021) is a collection of 8500 grade-school math word problems. The
MATH dataset (Hendryks et al. 2021) consists of 12,500 problems
taken from high school competitions. The GHOSTS dataset (Frieder et al., 2023)
has 703 problems taken
from advanced undergraduate and graduate math courses. 

Likewise, quite a few science benchmarks have been assembled, covering
a range of forms, fields, and levels. ScienceQA (Lu, 2022), is a collection
of 21,000 multimodal science questions, annotated with explanations. 
The Textbook Question Answering (TQA) dataset includes 1100 lessons and 26,000
multi-modal questions taken from middle-school science curricula.  The ARC
dataset (Clark et al. 2018) is a collection of 7800 multiple-choice 
science questions at a grade school. The BIG-bench collection (Srivastava
et al. 2022) contains 162 benchmarks that are tagged as dealing with
``math, logic, and code'' and 20 that are tagged as dealing with ``scientific
and technical understanding.''

\section{Discussion}
\label{secDiscussion}
Our test sets were too small and too haphazard in construction
to support statistically valid
conclusions, but they were suggestive of a number of conclusions. 

In our discussion here, we will refer to the problems in the ``Arbitrary''
test set as A.1-A.32; those in the ``Calculation-Free'' test set as B.1-B-53;
and those in the ``Motivated'' test set as C.1-C.20.

We feel confident that, over the kinds of problems tested, 
GPT-4 with either plug-in is significantly
stronger than GPT-4 by itself, or, almost certainly, than any AI that existed
a year ago.  (We do not have experimental evidence for the claim; we carried
out few tests on default GPT-4 and none at all on other systems.)
It is much stronger than Wolfram Alpha, whose natural language interface 
is very brittle.  These systems may well be the strongest AI systems for
these problem currently in existence.
However they are still far from reliable; they often output a wrong
answer or fail to output any answer.

The absolute numerical scores --- ``GPT-4+WA got 8.25/32 on the Arbitrary 
Numerical test set'' and so on --- are meaningless in themselves, since
the construction and choice of the problems was arbitrary and non-replicable.
Broadly speaking, though, we would guess\footnote{We have not run any human
subject tests.} that the overall performance of
these systems on these test sets was at the level of a middling
undergraduate student provided with the physical constants involved.
We would expect a capable math or physics major in their sophomore year
to get a good score, if they were provided with the physical
constants involved and could use a calculator.
However, the capacities and
weaknesses of these AI systems
do not align with a human student; the systems solve some 
problems that even strong students would
find challenging, such as the linear algebra problem C.14 , whereas they 
fail on some problems that even middling
high school students should find easy, such as the eclipse problems B.1-B.8.

It would seem, though our tests are too small to establish conclusively,
that the two plug-ins are incomparable in power; each can solve problems
that the other cannot. Neither seems decisively superior overall to the
other.

There is considerable room for improvement with the interfaces between GPT-4 
and the plug-ins, particularly with Wolfram Alpha. GPT-4 often struggles to
formulate a problem in a way that Wolfram Alpha can accept or that produces
useful output. For instance, in problem B.35, GPT-4 called Wolfram Alpha 40 
times unsuccessfully trying to get the date of the Hundred Years' War and 
then gave up (GPT-4 without plug-ins knows that date in any case.) In problem
C.14, a probabilistic calculation, the answer returned by Wolfram Alpha was
in the form of the exact rational number with a numerator of 1521 digits and a 
denominator of 1522 digits. GPT-4 was unable to estimate its value, and it
did not occur to GPT-4 to rephrase the request to Wolfram Alpha to return
the value in an intelligible form.  In problem A.14, which asked for an answer
in astronomical units, Wolfram Alpha returned the correct answer expressed
in meters, but then the system was unable to get that converted to astronomical
unites, despite making eight calls to Wolfram Alpha.

GPT-4 sometimes created useless calls to
the plug-ins, which asked for no useful information. For example, in problem
C.11, GPT carried out the following exchange with WA:

\begin{verbatim}
Request to Wolfram
{
  "input": "F = m*a; F"
}
Response from Wolfram
"a m"
\end{verbatim} 

In problem C.15, GPT-4 issued the following call to CI:

\begin{verbatim}
pythonCopy code
# Total weight of gold and uranium mined (in kg)
gold_weight_kg = 190_000_000  # kg
uranium_weight_kg = 2_500_000_000  # kg

gold_weight_kg, uranium_weight_kg
\end{verbatim}

That is, the plug-ins simply being asked to echo formulas or 
values that GPT-4 itself provided.

GPT-4 fails to take full advantage of the capacities of the plug-ins, 
The LLM often carries out calculations, retrieves formulas, and does symbolic
manipulations that the specialized plug-ins could carry out more reliably.
This sometimes leads to errors that should have been avoidable.
For instance in problem C.9, GPT4 made a mistake in a symbolic 
manipulation of a formula; either WA or {\tt sympy} can carry out such
operations flawlessly. In problem A.20, which involved computing compound 
interest over a number of years, the ``natural'' solution would be to write 
a small for loop to carry out the calculation, and indeed GPT4+WA did
exactly that; however, GPT4+CI attempted to carry out the loop purely in
GPT, without calling CI, which led to an erroneous answer. In the sorting 
problems B.33-B.39, GPT4  never called either plug-in to carry out the sorting;
it did the sorting itself, and not always correctly.

Overall, the systems are strongest on problems that can be solved by invoking
a single formula. They are often weak on problems that humans would tend to 
solve using spatial visualization, such as the eclipse problems B.1-B.8. 
They are often weak on problems that involve
combining several calculations of different kinds, such as the pendulum
problem A.12 or the chemical reaction problem A.24.  The systems sometimes have
trouble dealing with numbers that are very large or very small. The 
systems seem to have particular difficulty with the theory of special
relativity, when problems go beyond simply plugging in one of the standard
formulas (problems A.3 and C.11).

GPT-4 has some ability to detect when the answer returned by a plug-in is
nonsensical or physically meaningless, but not reliably. 
Its performance would be improved were it better able to diagnose
the reasons for the mistakes and to recover from them.

Over the course of these problems, GPT-4 does a fair amount of arithmetic
calculations over floating point numbers without calling either plug-in.
Occasionally it makes a mistake -- for instance, in problem B.20 it decided
that \$1449.46 is larger than \$1823.26 ---  
but much more frequently, it gets the
correct answer. This is quite remarkable in itself.

Both of these systems tend to use ``brute force'' methods which can involve
unnecessary computation. In particular, they almost always convert quantities
to standard units even when that conversion is unnecessary. For instance,
in problem C.11, both systems used the value for the speed of light, even 
though the problem can be solved taking $c=1$. In problem A.25, dealing with
radioactive decay, GPT4+WA brought in Avogadro's number, even though
the problem can be solved entirely at the level of masses and moles. On the
other hand, in problem A.5, it found a solution path more elegant than Davis'
and in problem A.25, its solution correctly incorporated a significant 
consideration that Davis had overlooked.

It seems likely that GPT4+CI and GPT4+WA are most useful 
when not relied on as
``oracles," but in ``interactive mode," with a human using them to speed
up problem-solving but also making multiple queries, checking that the
outputs look reasonable, and so on. We examined this option only
in two cases, but it deserves further study.

\subsection{Methodological observations}
We conclude with a couple of methodological observations. First: The methodology
we have used in these tests is, currently, a somewhat unusual one for AI
systems: We have personally constructed a small, idiosyncratic, test set, 
we manually examined 
the entire output on that set, we carried out a quite painstaking analysis,
and we have published the entire test set with hand-written explanations 
of the answers and the entire outputs of both systems, annotated
and lightly edited. 
This approach does not yield conclusions that to any degree support statistical 
analysis or statistical measures of confidence, and it would expensive to
scale it to a test set large enough to support that analysis.  
However, it does enable
the detailed analysis of the strengths and weaknesses of these systems
that we have given in section~\ref{secDiscussion} above.

Second: Our analysis relies heavily on the explanations generated by the 
AI systems in their output. In general, that kind of reliance can be
risky in analyzing the behavior of LLMs. In general an LLM generates
an answer to a question by finding the most likely sequence of tokens to 
fit the ``answer'' slot; it generates the explanation of the answer by
finding the most likely sequence of tokens to fit the ``explanation'' slot;
the explanation may have nothing to do with the way in which it actually 
generated
the answer. However, for many of the problems in our test set, that 
caution is mitigated. One can be reasonably confident that the explanation
is, in fact, related to the answer for two reasons:

\begin{itemize}
\item[1.] In problems where GPT-4 calls the plug-in and uses the answer 
supplied by the plug-in, the answer is clearly a consequence of the plug-in 
query.
\item[2.] In the numerical problems (except the handful, such as C.1, where
the answer can be directly Googled) the only way to arrive at the correct
answer is to carry out a correct calculation. In those problems where GPT
does not call the plug-ins, it is certainly {\em possible\/}
that the AIs are using some {\em other\/} calculation than the one that
it describes, but there is no reason to suppose that.
\end{itemize}

In the multiple-choice and true-false problems of the ``Calculation-Free'' 
test set, these considerations almost never apply --- even when GPT-4 calls
the plug-in, it is not to compute the final answer --- so there the
relation of the explanation to the answer is considerably more tenuous. Also,
the relation of explanations of things other than the numerical answer may
be questionable. For instance, if the LLM retrieves a particular formula
and gives a verbal explanation of why that formula is relevant, the
explanation may not be the actual reason that the LLM retrieved the formula.
The LLM may well have found the formula and explanation separately
by, in effect, matching the wording of the problem to similar problems 
in the training
set, and retrieving the formula and explanation separately from those. For
instance, it would be possible for an LLM to retrieve a formula from one
source in the training set that discusses the problem and to retrieve
an unrelated explanation from another source discussing the same problem;
we did not observe any such behavior in our tests, but, as far as we know,
nothing in LLM technology makes that impossible.

\subsection*{References}
Kurt Cobbe et al. (2021). Training verifiers to solve math word
problems. \\
\url{https://arxiv.org/abs/2110.14168}

Simon Frieder et al. (2023). Mathematical Capabilities of ChatGPT.
\url{https://arxiv.org/abs/2301.13867}

Dan Hendryks et al. (2021). Measuring mathematical problem solving with 
the MATH dataset. 
\url{https://arxiv.org/abs/2103.03874}

Pan Lu et al. (2022). Learn to explain: Multimodal reasoning via 
thought chains for science question answering. 
{\em Advances in Neural Information Processing Systems,} {\bf 35}:2507–2521.

Todor Mihaylov, Peter Clark, Tushar Khot, and Ashish Sabharwal (2018). 
Can a suit of armor conduct electricity?
a new dataset for open book question answering. 
{\em Proceedings of the 2018 Conference on Empirical
Methods in Natural Language Processing (EMNLP).}

Swaroop Mishra et al. 2022. {\sc L\=ila}: A unified benchmark for 
mathematical reasoning. \\
\url{https://arxiv.org/abs/2210.17517}

OpenAI (2023a). GPT4. March 14, 2023.
\url{https://openai.com/research/gpt-4}

OpenAI (2023b). GPT4 Technical Report. March 16, 2023.
\url{https://arxiv.org/abs/2303.08774}

OpenAI (2023c). ChatGPT plugins. March 23, 2023.
\url{https://openai.com/blog/chatgpt-plugins}

Arkil Patel, Satwik Bhattamishra, and Navin Goyal. Are NLP Models 
really able to Solve Simple Math Word Problems?. 
\url{https://arxiv.org/abs/2103.07191}

Aarohi Srivastava et al. 2022.
Beyond the imitation game: Quantifying and extrapolating
the capacities of language models.
\url{https://arxiv.org/abs/2206.04615}

Xiaoxuan Wang et al. (2023). SciBench: Evaluating college-level scientific 
problem-solving abilities of large language models.
\url{https://arxiv.org/abs/2307.10635}

Steven Wolfram (2023). ChatGPT gets its ``Wolfram superpowers''!. March 23,
2023. \\
\url{https://writings.stephenwolfram.com/2023/03/chatgpt-gets-its-wolfram-superpowers/}

Zicheng Ye et al. (2023).
Improving the Gilbert-Varshamov Bound by Graph Spectral Method.
{\em CSIAM Transactions on Applied Mathematics,} {\bf  4}(1): 1–12 \\
\url{https://global-sci.com/article/90844/improving-the-gilbert-varshamov-bound-by-graph-spectral-method}

\subsection*{Appendix A: The ``Arbitrary Numerical'' test set}
\label{secNumerical}

The Matlab code included here was the code that we ourselves wrote to carry 
out the calculations. It is included here
to facilitate checking the correctness of our answers.

\begin{enumerate}
\item 
Object A is a spherical shell of gold, 1 meter in radius, 10 cm thick.
Object B is a spherical shell of copper, 10 cm in radius, 5 cm thick.
They are placed so that there is an 8 cm gap between them. What is the net
gravitational force they exert on each other?

{\bf Answer:} 3.4486 dyne

{\bf Matlab}
\begin{verbatim}
% By a theorem of Newton's, the gravitational force exerted by spherical shell
% is the same as if all the mass were concentrated at the center.
GoldDensity = 19.3; % gm/cm^3 --  Wikipedia.
GoldVolume = (4/3) * pi * (100^3-90^3); %cm^3
GoldMass = GoldDensity * GoldVolume; % gm
CuDensity = 8.96    % gm/cm^3 --- Wikipedia
CuVolume = (4/3) * pi  * (10^3-5^3);  % cm^3
CuMass = CuDensity * CuVolume;
Distance = 100+10+8; %cm
GravConst = 6.674 * 10^(-8); % dyne cm^2/gm^2 -- Wikipedia
Force = GravConst * GoldMass * CuMass / Distance^2
\end{verbatim}

\item Let b, c, d be points at the center of Buenos Aires, Cincinnati, and Delhi,
respectively. Consider the plane P that contains b, c, and d (it cuts through
the earth). What is the area in square kilometers of the triangle b,c,d
lying in P?

{\bf Answer:} 41,111,600 $\mbox{km}^{2}$.

{\bf Matlab}
\begin{verbatim}
% Convert latitude and longitude to a unit 3D vector
function v=LatAndLongToUnit(LatDeg, LatMin, LatS, LongDeg, LongMin, LongS)
Theta = LatS*(LatDeg+(LatMin/60))*(pi/180);
Phi = LongS*(LongDeg+(LongMin/60))*(pi/180);
v = [cos(Theta)*cos(Phi), cos(Theta)*sin(Phi), sin(Theta)];
end

b=LatAndLongToUnit(34, 36, -1, 58, 22, -1)
c=LatAndLongToUnit(39, 6, 1, 84, 30, -1)
d=LatAndLongToUnit(28, 37, 1, 77, 14, 1)
bc=c-b
bd=d-b
cd=d-c
EarthRadius=6371;                          % The magnitude of the cross product 
answer = norm(cross(bc,bd))*EarthRadius^2/2  % is the area of the parallelogram
x=norm(bc)
y=norm(bd)
z=norm(cd)
s=(x+y+z)/2
check = sqrt(s*(s-x)*(s-y)*(s-z))*EarthRadius^2  % Heron's formula
\end{verbatim}

\item X is a regular tetrahedron of zinc with a side length of 2 moving at 100
m/sec. Y is a molecule of NaCl that is moving with speed $v$. X and Y have 
equal kinetic energy. What is $1-(v/c)$, where $c$ is the speed of light?

{\bf Answer:} $8.41 \cdot 10^{-21}$

{\bf Matlab:}
\begin{verbatim}
ZincVolume=2^3/(6*sqrt(2)) % Wolfram Alpha
ZincDensity = 7.14 % gm/cm^3. Wikipedia
ZincMass = ZincVolume*ZincDensity
VZinc = 100/(3*10^8) % Speed of Zinc in units where C=1
ZincKE = ZincMass*VZinc^2
NaClMass = 9.7*10^(-23) % gm. Wolfram Alpha
LorenzFactor = ZincKE/NaClMass
Q = 1/LorenzFactor^2 % 1-(v/c)^2
Answer = Q/2
\end{verbatim}

Note: The above was the question, answer, and Matlab code that we published in
August 2023. We observed in September 2024 that the question omitted the measure of 
length in describing the zinc prism, and that in the fifth line of our Matlab code,
the right side should be divided by 2. The correct answer to this question is 
$3.364 \cdot 10^{-20}$

\item Viewed from Vega, what is the angle between Sirius and the Sun?

{\bf Answer:} 0.0972 radians

{\bf Matlab:}
\begin{verbatim}
AngleSEV = 2.755 % radians. Wolfram Alpha
DistES = 8.597 % light years. Wolfram Alpha
DistSV = 33.41 % light years. Wolfram Alpha
AngleEVS = asin(sin(AngleSEV)*DistES/DistSV)
\end{verbatim}

\item You have an empty cylindrical open container whose inside has a diameter of 
8 centimeters and a height of 20 centimeters. and a pitcher with 200 ccs of water.
You first empty the pitcher into the cylinder, then
put a solid rock cube, 4 cm on a side,  into the container so that it is
sitting flush against the
bottom of the container. What is the height of the water in the container?

{\bf Answer:} 5.2521 cm

{\bf Matlab:}
\begin{verbatim}
AreaOfCyl = pi*4^2
AreaAtBottom = AreaOfCyl-4^2
VolumeAtBottom = 4*AreaAtBottom
VolumeAtTop = 200 - VolumeAtBottom
HeightAtTop = VolumeAtTop/AreaOfCyl
Height = 4+HeightAtTop
\end{verbatim}

\item There are six small spherical masses placed at various positions in
space. Measured in a coordinate system with a 1 meter unit length: \\
A has mass 2 kg and is at $\la 2,0,0 \ra$; \\
B has mass 3 kg and is at $\la 0,2,0 \ra$; \\
C has mass 1 kg and is at $\la 0,0,2 \ra$; \\
D has mass 4 kg and is at $\la 1,1,1 \ra$; \\
E has mass 5 kg and is at $\la 1,0,1 \ra$; \\
F has mass 2 kg and is at $\la 0,1,1 \ra$. \\
What is the instantaneous acceleration of F under the gravitational attraction
of A,B,C,D,E? Express your answer as a vector. 

{\bf Answer:} $\la 0.403,-0.080,-0.056\ra *10^{-9} \mbox{ m/sec}^2$ 

{\bf Matlab:}
\begin{verbatim}
% Gravitational force exerted on mass ma at point pa by mass
% mb at point pb, taking the gravitational constant to be 1.

function f = GravForce(ma,pa,mb,pb)
    v = pb-pa;
    r = norm(v);
    u = v/norm(v);
    f = ma*mb*u/(r^2)
end

g1 = GravForce(2,[0,1,1],2,[2,0,0]) + ...
GravForce(2,[0,1,1],3,[0,2,0]) + ...
GravForce(2,[0,1,1],1,[0,0,2]) + ...
GravForce(2,[0,1,1],4,[1,1,1]) + ...
GravForce(2,[0,1,1],5,[1,0,1])
GravConst = 6.674*10^(-11) % N*kg^2/m^2 -- Wikipedia
acc = g1*GravConst/2
\end{verbatim}

\item How many total eclipses of the moon were there between Jules Verne's 
death and Neil Armstrong's moon landing? An exact integer value is required.

{\bf Answer:} 52.

Jules Verne died March 24, 1905. \\
Neil Armstrong landed on the moon July 20, 1969 \\
There were 52 total lunar eclipses in between: \\
02-09-06, 08-04-06, 06-04-09, 11-27-09, 05-24-10, 11-17-10, \\
03-22-13, 09-15-13, 01-08-17, 07-04-17, 12-28-17, 05-03-20, \\
10-27-20, 04-22-21, 02-20-24, 08-14-24, 06-15-27, 12-08-27, \\
06-03-28, 11-27-28, 04-02-31, 09-26-31, 01-19-35, 07-16-35, \\
01-08-36, 05-14-38, 11-07-38, 05-03-39, 03-03-42, 08-26-42, \\
12-19-45, 06-14-46, 12-08-46, 04-13-49, 10-07-49, 04-02-50 \\
09-26-50, 01-29-53, 07-25-53, 01-19-54, 11-18-56, 05-13-57, \\
11-07-57, 03-13-60, 09-05-60, 12-30-63, 06-25-64, 12-19-64, \\
04-24-67, 10-18-67, 04-13-68, 10-06-68 \\
Source: Wikipedia article, ``List of lunar eclipses in the twentieth century.''

\item A quantity of chlorine gas is in a right prism whose base is a triangle with
sides 5 cm, 7 cm, and 4 cm and whose altitude is 8 cm. The temperature is 
the freezing point of mercury, and the pressure is 2 atmospheres. What is the 
mass of the chlorine?

{\bf Answer:} 0.5781 gm

{\bf Matlab:}
\begin{verbatim}
s=(5+7+4)/2
Area = sqrt(s*(s-5)*(s-7)*(s-4))  % Heron's formula
VolumeCC = Area * 8
Liters = VolumeCC / 1000
Temp = 273.15 -38.83  % Wikipedia
IdealGasConst = 0.082057 % L*atm*K^{-1}*mol^{-1}. Wikipedia
Moles = Liters*2/(Temp*IdealGasConst) 
MassPerMoleCl2 = 70.9 % gm. Wolfram Alpha.
Mass = Moles * MassPerMoleCl2
\end{verbatim}

\item A train has two whistles, one at middle C and one at the F above middle C.
The train is driving past a station without stopping or changing speed. 
On approaching the station, it blows
the low whistle; when it passes the station, it switches to the high whistle.
However, to the people standing at the station, it sounds like the whistle
dropped by a whole tone. How fast is the train moving?

{\bf Answer:} 68.4 m/s

{\bf Explanation:}
A half-tone in music is a factor of $2^{1/12}$. The interval
from C to F is 5 half-tones. So the effect is that a note as the train approaches is dropped by 7 half tones or a factor of
$2^{7/12}$.
Let $s$ be the speed of sound, then $(s+v)/(s-v) = 2^{7/12}$.
Solving for $v$ we get
\[ v = s \cdot \frac{2^{7/12}-1}{2^{7/12}+1} \]
The speed of sound is 343 m/s so the train
is moving at 68.4 m/s. (One can work this out by hand to an accuracy of 1\% using
the fact that a fifth in music -- 7 half tones -- is 1.5 in the natural musical
scale.) 

\item 
A physical process generates photons whose energies follow a random distribution of the following form: For positive energy e, the probability
density at e is proportional to the value of e in a Gaussian distribution
with mean 2 Ev and standard deviation 0.01 Ev. The probability of a negative value is zero. What is the expected value of the wavelength of a
photon produced by this process? (Give the mathematical answer, assuming that the above description is exact, and assuming the standard relation between energy and wavelength in a photo. The answer is not physically plausible.)

{\bf Answer:} Infinite.

{\bf Explanation:} The wavelength of a photon with energy $e$ is $hc/e$. The
mean wavelength of a photon in the beam is therefore

\[ \int_{e=0}^{\infty} N_{2,0.01}(e) \cdot \frac{hc}{e} \: de \]

In the limit as $e \goesto 0^{+}$, the quantity $N_{2,0.01}(e)$, though 
extremely tiny, is still a positive quantity, so the integral diverges
at 0.

\item The wavelengths of the photons in a beam of light are uniformly 
distributed across the range of visible (to humans) light.
What is the mean energy of a photon in electron volts?

{\bf Answer:} 2.3688 eV.

{\bf Explanation:} The range of visible light is around 3800 to 7000 Angstroms.
A photon with wavelength $\lambda$ has energy $hc/\lambda$. Thus, if $\lambda$
is uniformly distributed between 3800 and 7000, the mean energy is

\[ \int_{\lambda=3800}^{7000} (1/3200) (hc/\lambda) \: d\lambda =
(hc/3200) \log_{e} \lambda |^{7000}_{3800} \]

{\bf Matlab:}
\begin{verbatim}
Planck = 4.136 * 10^(-15) % eV*sec
SpeedOfLight = 3*10^8 % m/sec
Meter2Ang = 10^(10) 
SpeedOfLight * Meter2Ang * Planck / 3200 * log(7000/3800) 
\end{verbatim}

\item A pendulum is hanging on a 2 meter cord attached to the ceiling 3 
meters above the floor. It is brought to a position 25 degrees from the 
vertical and released. It swings past the bottom and the cord is cut when it is
10 degrees from the vertical on the far side. Then the bob flies through the
air and hits the ground. What is the distance from the point where the bob
is released to the point where it hit the ground?

{\bf Answer:} 2.3606 meters

\begin{verbatim}
% Take the point of attachment of the pendulum as the origin
g = 9.8  % Acceleration of terrestrial gravity in m/sec^2
d2r = pi/180 % Degrees to radians
s25 = sin(25*d2r) % Handy abbreviations
c25 = cos(25*d2r)
s10 = sin(10*d2r)
c10 = cos(10*d2r)
pr = [-2 * s25, -2*c25]  % release point
pc = [2 * s10,  -2*c10]   % cut point
hd = 2 * (c10 - c25)  % height difference between the cutpont and the release point.
s = sqrt(2*g*hd) % speed of the bob at release time
vc = s * [c10,s10]; % velocity of bob when cut
tca = vc(2)/g      % time between cut and apex of flight
hca = (g*tca^2)/2  % height difference from cut point to apex 
ha = -2*c10 + hca % height of apex of flight
haf = 3 + ha      % height difference between apex and floor
taf = sqrt(2*haf/g) % time from apex to floor
pf = [pc(1) + (tca+taf)*vc(1),-3]
answer = norm(pf-pr)
\end{verbatim}

\item An spherical asteroid 500 km in diameter
travels in an essentially perfect circular orbit of radius 
2.4 astronomical units. On January 4, 2023, the earth was at perihelion, and
as it happens, the asteroid was in perfect opposition. What was the solid angle
of the asteroid in the sky as seen by an earth observer? 

{\bf Answer:} $4.35 \cdot 10^{-12}$

{\bf Matlab:}
\begin{verbatim}
perihelion = 1.471 * 10^8 % km. Wikipedia
au = 1.498 * 10^8 % km. Wikipedia
asteroid = 2.4*au
distance = asteroid - perihelion
radiusAngle = 250/distance
solidAngle = pi*radiusAngle^2
\end{verbatim}

\item Two twin stars, one of 5 solar masses, the other of 10 solar masses, 
orbit each other in circular orbits in a period of 4 earth years. How far apart
are they, in astronomical units? 

{\bf Answer:} 6.21 Au.

{\bf Explanation:} \\
Let $L$ be the distance between them in AUs. They each rotate around in a circle
centered at their center of gravity,
which is $2L/3$ from the small star and $L/3$ from the large star. \\
Let $M = 5$ be the mass of the smaller star in solar masses.\\
Let $Q$ be the ratio of the centripetal acceleration of the 
large star to the centripetal
acceleration of the earth in its orbit. \\
Comparing the gravitational acceleration, $Q = M/L^{2} = 5/L^2$
Comparing the centripetal acceleration, the star is moving on a path of 
radius $L/3$ Au at an 
angular velocity 1/4 of earth. Hence $Q=L/48$. 
So $L^{3} = 5 \cdot 48$, so $L$=6.21 Au.

{\bf Matlab:} Alternative solution using a generalized form of Kepler's third
law: \\
$a^{3}/T^{2} = G(m_{1}+m_{2})/4 \pi^{2}$.
\begin{verbatim}
Year = 3.154*10^7 %sec
Grav = 6.6743*10^(-11) % N*m^2/kg^2
SolarMass = 1.99*10^(30) % kg
M1 = 10*SolarMass
M2 = 5*SolarMass
Period = 4*Year
DistInMeter = (Grav*(M1+M2)*Period^2/(4*pi^2))^(1/3)
DistInAU = DistInMeter/(1.496*10^11)
\end{verbatim}

\item Draw a circle, on the earth's surface, going through Cairo, Peking, and 
Moscow.  Let S be the area of the part of the earth's surface inside the circle 
and let P be the area of the circle in the plane of the circle. What is S/P?.

{\bf Answer:}  1.108.

{\bf Matlab:}
\begin{verbatim}
% Since we are computing a ratio, we take the earth's radius to be 1
c = LatAndLongToUnit(30,2,1,31,14,1) % Unit vector corresponding to Cairo
p = LatAndLongToUnit(39,54,1, 116,24,1) % Peking
m = LatAndLongToUnit(55,45,1,37,37,1) % Moscow
a = zeros(3);  % coefficient matrix
q = zeros(3,1); % constant terms
% Equation of the perpendicular bisector plane of line cp is dot(x,p-c) = 0
% (Note that the origin is always in the plane.)
a(1,:) = p-c
% Likewise perpendicular bisector of cm
a(2,:) = m-c
a(3,:) = cross(p-c,m-c) % normal to plane of three cities
q(3,1) = dot(c,a(3,:))
f = a\q  % circumcenter is intersection of these 3 planes
f=f';
r = norm(f-c) % radius of circumcenter
norm(m-f)  % check for correctness
norm(p-f)
PlanarArea = pi*r^2  % area of planar circle
SphereArea = 2*pi*(1-norm(f)) % Formula from Wikipedia "Spherical sector"
Answer = SphereArea/PlanarArea
\end{verbatim}

\item  Assume that the probability of having any particular isotope of a chemical
follows their frequency on earth. What is the probability that a randomly
constructed molecule of glucose will have 
4 atoms of $^{12}\mbox{C}$,
2 atoms of $^{13}\mbox{C}$,
11 atoms of $^{1}\mbox{H}$,
1 atom of $^{2}\mbox{H}$,
3 atoms of $^{18}\mbox{O}$,
3 atoms of $^{16}\mbox{O}$.

{\bf Answer:} $4.797 \cdot 10^{-13}$

{\bf Matlab:}
\begin{verbatim}
nchoosek(6,4) * .0106^2 *0.989^4 * ...
nchoosek(12,1) * .99985^11 *0.000145 * ...
nchoosek(6,3) * 0.998^3 * 0.00205^3
Answer: 4.797 * 10^-13
\end{verbatim}

\item Two $^{31}K$ phosphorus nuclei, with no electrons, are isolated in
space, with coordinates $\la$0,0,0$\ra$ and $\la$10,10,10$\ra$ in a coordinate
system whose unit length is 1 Angstrom. What is the instantaneous acceleration
(a vector with unit length of Angstrom/sec$^{2}$)
of the nucleus at the origin due to the electrostatic force?

{\bf Answer:} $\la -1.9420 \cdot 10^{27},  -1.9420 \cdot 10^{27},  
-1.9420 \cdot 10^{27} \ra$

{\bf Matlab}
\begin{verbatim}
Mass = 5.1433*10^(-26) % kg
Charge = 15*1.602*10^(-19) % coulomb
CoulombConst = 8.988 * 10^9 % Newton * m^{2}/C^{2}
Angstrom2Meter = 10^(-10)
Distance = 10*sqrt(3)*Angstrom2Meter
Force = CoulombConst*Charge^(2)/Distance^(2) % Newtons
Acceleration = -Force/(Mass*Angstrom2Meter) * [1,1,1]/sqrt(3)
\end{verbatim}

Note: The above was the question, answer, and Matlab code that we published in
August 2023. We noticed in September 2024 that the question was mistated; the
symbol for phosphorus is P, not K. Our calculation was right on the assumption that
phosphorus was intended.

\item An irregular (house-shaped) pentagon has vertices numbered 1 through 5 in
order. The pentagon has right angles at vertices 1, 3, and 4, and 135-degree
angles at 2 and 5. Side 2-3 and 4-5 have length 1 and side 3-4 has length 2.
The pentagon is placed on a planar coordinate system so that the numbering of
the vertices is in clockwise order, vertex 3 is at the origin, and vertex 5
is on the positive y-axis. What are the coordinates of vertex 1?

{\bf Answer:} $\la 1.5269, 2.3446 \ra$

{\bf Matlab:}
\begin{verbatim}
% Problem 18
% Consider the pentagon in a standard orientation where vertex 3 is at the origin and
% vertex 4 is on the negative x-axis. Compute the rotation necessary to place 5 above
% 3, and then apply that rotation to 1.

theta = atan2(1,-2)-(pi/2)
standard1 = [-sqrt(2);1+sqrt(2)]
answer = [cos(theta), sin(theta); -sin(theta), cos(theta)]*standard1 
standard5 = [-2;1]
check = [cos(theta), sin(theta); -sin(theta), cos(theta)]*standard5 % check 
\end{verbatim}

\item Consider a cube with unit length sides, where the vertices of one
face are numbered A..D in counterclockwise order, as viewed from the center
of the cube;  the vertices of the opposite face are named E to H; and there are
edges AE, BF, CG, and DH.
Rotate the cube so that vertex A is at the origin,
vertex G is on the positive z axis, and vertex B is in the x-z plane, with
positive x coordinate.
What are the coordinates of vertex E?

{\bf Answer:} $\la -0.4082, -0.7071, .5774 \ra = \la -\sqrt{1/6}, 
-\sqrt{1/2}, \sqrt{1/3} \ra$ 

{\bf Explanation:} 
Consider the cube in a standard position where $\vec{a} = \vec{0}$,
$\vec{b}=\hat{i}$, $\vec{d}=\hat{j}$. $\vec{e}=\hat{k}$; thus 
$\vec{g} = \la 1,,1,1 \ra$. Since $\vec{a}$ is at the origin after the rotation,
this is a pure rotation around the origin; it is therefore equivalent to
multiplication by an orthogonal matrix $m$ with determinant 1. 
The dot product $m(2,:)$ with any vector $v$ is the $y$-coordinate of the
rotated position of $v$. In particular since after the rotation both $\vec{b}$
and $\vec{g}$ have y-coordinate zero, $m(2,:)$ must be orthogonal to both; 
hence, it is their cross-product, normalized. Since the $x$ coordinate of 
the rotated place of $g$ is 0, and since $m$ is orthogonal, $m(1,;)$ is
orthogonal to both $\vec{g}$ and $m(2,:)$; hence, it is the normalized
form of their cross product. Since $m$ is orthogonal, $m(3,:)$ is the
cross product of $m(1,:)$ and $m(2,:)$. All that
remains is to make sure the signs are OK, to check, and to compute
$m \cdot \vec{e}$.

{\bf Matlab}

\begin{verbatim}
% Problem 19

b = [1,0,0]
g = [1,1,1]
e = [0,0,1]
m = zeros(3);
u = cross(b,g) % both b and g get mapped to a vector with y component = 0
m(2,:) = u / norm(u)
v = cross(m(2,:),g) 
m(1,:) = sign(dot(v,b))*v/norm(v)
m(3,:) = cross(m(1,:),m(2,:))
if (dot(m(3,:),g) < 0)
   m(2,:) = -m(2,:)
   m(3,:) = -m(3,:)
end
m*m' %Check: rotation matrix?
det(m)
m*b'  %Check: b mapped into xz plane with positive x?
m*g'  %Check: g mapped into positive z axis?
answer = m*e'
\end{verbatim}

\item 
Joe and Jim each have a bank account which they started on January 1,
2000. Joe started his account with with \$1000; Jim started his with \$900.
Joe's account pays 5\% every December 31, and he
adds an additional \$500 every January 1. Jim's account pays 10\% every
December 31. When will Jim's account have more money in it than Joe's?

{\bf Answer:} 2053. (More precisely, December 31, 2052).

{\bf Matlab:}
\begin{verbatim}
joe = 1000
jim = 900
count = 0
while (jim <= joe)
  count = count+1
  jim = jim*1.10
  joe = joe*1.05 + 500
end
2000+count
\end{verbatim}

\item  Two points {\bf p} and {\bf q} are independently randomly
chosen following
a uniform distribution in a three-dimensional
sphere of radius $10^{7}$. With a relative error of
less than 1\%, give an estimate with an error no greater than 1\%
that the Euclidean distance from
{\bf p} to {\bf q} is less than 1?

{\bf Answer:} $10^{-21}$.

{\bf Explanation:} For any point {\bf p} not close to the boundary of the
large sphere, {\bf q} will be within distance 1 of {\bf p} if and only if
it is inside the sphere radius 1 around {\bf p}. That breaks down if {\bf p}
is within 1 of the boundary of the large sphere, but the probability of that
is tiny (around $4 \pi \cdot 10^{-7}$). So the probability that {\tt q} is inside
the small sphere given that it is in the large sphere is the ratio of the volumes
of the spheres = $10^{-21}$.

\item A point {\bf p} is chosen at random within the 100-dimensional box
${\bf B}=[0,100]^{100}$ following a uniform distribution. 
What is the probability
that the Euclidean distance from {\bf p} to the boundary of $\bf B$ 
is less than 1?

{\bf Answer:} $1-0.98^{100} \approx 1-1/e^{2} = 0.8647$

{\bf Explanation:} The distance from {\bf p} to the boundary is
$\min_{i=1..k} \min(p_{k}, 100-p_{k})$ That is, the distance is more than 1 
if and only if ${\bf p} \in [1,99]^{100}$. So the probability that the distance
is less than one is $(100^{100}-98^{100})/100^{100}$.

\item A point {\bf p} is chosen at random within the 100-dimensional box
${\bf B}=[0,1]^{100}$ following a uniform distribution. What is the mean
value of the Euclidean distance from {\bf p} to the boundary of {\bf B}?

{\bf Answer:} 1/202.

{\bf Explanation:} Pick one particular face of the hypercube; for convenience
the lower face in the x-dimension ${\bf F} = \{ 0\} \times [0,100]^{99}$.
Let ${\bf c} =  \la 0.5, 0.5, \ldots 0.5 \ra$ 
be the center of {\bf B}. Let {\bf P}
be the hyperpyramid consisting of the convex hull of {\bf F} $\cup$ \{ {\bf c}
\}. It is obvious, and easy to show, that the points in {\bf B} whose 
closest boundary point is {\bf F} are exactly those in {\bf P} and that the
distance of a point in {\bf P} to {\bf F} is just its x-coordinate. Moreover,
since {\bf B} can be partitions into such pyramids, one for each face of 
{\bf B} and they all have identical distributions of distances to the boundary,
the mean distance to the boundary is the same in {\bf P}. Therefore the mean 
distance for a point in {\bf G} to the boundary of {\bf B}
is the same as mean distance for the distance from a point in {\bf P}
to {\tt F} which is just the mean x-coordinate of a point in {\bf P}.

The rest is just calculus. The cross-section of {\bf P} at x-coordinate $x$
is the box $[x,1-x]^{99}$ of 99-d volume $(1-2x)^{99}$. We could integrate
this to compute the volume of {\bf P} but it's simpler to note that 
there is one such pyramid for each face of {\bf B}; there are 200 such faces;
and their union of the pyramids is all of {\bf B} which has volume 1. So the
volume of {\bf P} is 1/200. So the probability density of a uniform choice
in {\bf P} as a function of {\tt x} is $200  (1-2x)^{99}$.

The expected value of $x$ in {\bf P} is therefore

\[   \int_{x=0}^{0.5} 200 (1-2x)^{99} x dx =
200 \int_{u=0}^{1} u^{99}) \cdot ((1-u)/2) \cdot  (du/2) =
(200/4) \int_{u=0}^{1} u^{99}-u^{100} du = \]
\[ (200/4) \cdot (u^{100}/100 - u^{101}/101) |^{1}_{0} = 1/202. 
\]

\item A pound of graphite is inside a closed cylindrical container with a radius
of 50 cm and a height of 200 cm, filled with air at room
temperature  ($22^{\circ}$ C) and atmospheric pressure. The graphite
reacts with the oxygen, producing equal masses of carbon monoxide and
carbon dioxide until one of the reactant chemicals is exhausted. How
many moles of each chemical is in the container at the end?

{\bf Answer:} No oxygen, 11.86 moles of carbon monoxide, 7.61 moles of
carbon dioxide, 18.2 moles of carbon.

{\bf Matlab:}
\begin{verbatim}
CAtomMass = 12.011
OAtomMass = 16
COMolecMass = CAtomMass+OAtomMass
CO2MolecMass = CAtomMass+2*OAtomMass
MolesOfCarbon = 453.6/CAtomMass % Wikipedia
Temp = 273.15 + 22  % Wikipedia
Liters = pi*200*50^2/1000 % Volume in liters
IdealGasConst = 0.082057 % l*atm*K^{-1}*mol^{-1}. Wikipedia
MolesOfAir = Liters/(Temp*IdealGasConst)
MolesOfO2 = MolesOfAir * .2095
MoleFracOfCO = CO2MolecMass/(COMolecMass+CO2MolecMass)% Molar fraction of CO in product 
A = 2*MolesOfO2/(2-MoleFracOfCO)
MolesOfCO = MoleFracOfCO * A
MolesOfCO2 = (1-MoleFracOfCO) * A
MolesOfCO*COMolecMass - MolesOfCO2*CO2MolecMass %check
MolesOfCO + 2*MolesOfCO2 - 2*MolesOfO2  %check
EndMolesOfCarbon = MolesOfCarbon - (MolesOfCO+MolesOfCO2)
\end{verbatim}

\item Suppose that you have a gram of pure radium 223. What is the mass of
the helium that that will generate via fission in a week? \\
{\bf Answer:} 0.0248 gm

{\bf Explanation:} Radium 223 undergoes a series of nuclear reactions: \\
Radium 233 $\goesto$ Radon 219 + alpha \\
Radon 219 $\goesto$ Polonium + alpha \\
Polonium 215 $\goesto$ Lead 211 + alpha \\
Lead 211 $\goesto$ Bismuth 211 + beta \\
Bismuth 211 $\goesto$ Thallium 207 + alpha \\
Thallium 207 $\goesto$ Lead 207 + beta \\
Thus each atom of radium generates 4 alpha particles.

The first reaction has a half-life of 11.43 days; the others have half-lifes ranging
from fractions of a second to a few minutes.

{\bf Matlab:}
\begin{verbatim}
radiumMass = 223
heliumMass = 4
halfLife = 11.43 % Half life of radium 223 in days
f = (1/2)^(7/halfLife)
answer = (1-f)*4*heliumMass/radiumMass
\end{verbatim}

\item 
A small school has 100 students and three extramural activities: the basketball
team, the chess club, and the drama society. There are 60 students in the band,
48 in the chess club, and 55 in the drama club. There are 18 students who
do both band and chess, 28 who do both band and drama, 31 who do both chess
and drama, and 8 students who are in all three groups.
What is the probability that a student is not in the band given
that they are either in the drama club or the chess club but not both? \\

{\bf Answer:} 11/41

{\bf Explanation:} It is easily determined from the above information that
$BCD = 8$.
$BC\bar{D} = 10$
$B\bar{C}D = 20$.
$B\bar{C}\bar{D} = 22$.
$\bar{B}CD = 23$.
$\bar{B}C\bar{D} = 7$.
$\bar{B}\bar{C}D = 4$.
$\bar{B}\bar{C}\bar{D} = 6$.
The above answer is immediate.

\item You draw 20 cards from a deck of cards labelled 1..60. To your surprise,
the cards can be arranged as an arithmetic sequence. What is the probability
that the successive difference is 3?
Give an exact answer as a rational number.

{\bf Answer:} 3/66 = 1/22.

{\bf Explanation:}
There are 41 sequences of difference 1 (starting with every
value from 1 to 41); 22 sequences with difference 2 (starting from 1 to 22);
and 3 with difference 3 (starting points 1, 2, and 3).

\item You draw 5 cards from a deck of cards labelled 1 ... 200. To your
surprise the cards can be arranged as a geometric series. What is the
probability that the smallest number is less than or equal to 7?
(Note that the ratio need not be an integer.)

{\bf Answer:} 9/16

{\bf Explanation:}
The possible increasing ratios are 2, 3, and 3/2.

If the ratio is 2, the sequence has the form
a, 2a, 4a, 8a, 16a. a can be any integer from 1 to 12.
In 7 of these, a $\leq$ 7.

If the ratio is 3, the sequence has the form
a, 3a, 9a, 27a, 81a. a is either 1 or 2.
In both of these, a $\leq$ 7.

If the ratio is 3/2, the sequence has the form
16a, 24a, 36a, 54a, 81a, and a can be either 1 or 2. In both of them
the smallest number is greater than 7.

So the probability is 9/16.

\item In this question and the following three,
assume that the satellite is moving in a closed orbit around the Earth
and that the only influence on the satellite's motion  is the
Earth's gravity. Assume that the Earth is a perfect sphere.
Ignore the revolution of the Earth around the sun,
but do not ignore the rotation of the Earth around its axis.

A satellite in a circular geosynchronous orbit remains directly above the
point on the earth's equator $0^{\circ}$ N, $50^{\circ}$ E. What is the
straight line distance from the satellite to Houston?

{\bf Answer:} 46,934 km. 

{\bf Matlab:}
\begin{verbatim}
% Method 1: Compute the positions of Houston and the satellite
% as 3 dimensional vectors
r = 6378.1 % km. Earth's radius
q = 35786 % height of geosynchronous orbit over earth's surface
rs = r+q % distance from center of earth to satellite
deg2rad = pi/180;  % degrees to radians
hlat = 29.75*deg2rad    % Houston latitude and longitude
hlong = -95.38*deg2rad
uh = [cos(hlat)*cos(hlong), cos(hlat)*sin(hlong),sin(hlat)] % unit vector Houston
h = r*uh % Houston location as vector
slong = 50*deg2rad
us = [cos(slong),sin(slong),0] 
s = rs*us % Satellite location as vector
answer1 = norm(s-h)

% Method 2 Apply law of cosines to triangle Houston - center of earth- satellite
cosTheta = dot(us,uh) % cosine of angle at center of earth
sqrt(r^2 + rs^2-2*cosTheta*r*rs)
\end{verbatim}

\item A satellite in a circular geosynchronous orbit passes directly above the
North and South Poles. When it crosses the North Pole, its velocity is in the
plane of the $0^{\circ}$ circle of longitude. At what longitude 
does it pass directly above a point in the Tropic of Cancer?
(An answer will be marked correct if it is within 3 degrees of the correct
answer.)

{\bf Answer:} $66.56^{\circ}$ W.

{\bf Explanation:} The period of a geosychronous orbit is equal to a sidereal
day. In that time, the satellite executes a $360^{\circ}$ revolution and the
earth executes a $360^{\circ}$ rotation, so their angular velocities are equal.
So the longitude is equal to the difference in latitude between the North Pole
and the Tropic of Cancer. Since the Tropic of Cancer is at 23.44 degrees, the
difference is 66.56 degrees.

\item A satellite orbits the earth in a circular orbit.
It completes an orbit every 14 hours 40 minutes. How high is
it above the Earth's surface?

{\bf Answer:} 24,042 km

{\bf Matlab:}
\begin{verbatim}
%  By Kepler's 3rd law, R^3 is proportional to T^{2}, so R is proportional to T^(2/3)
r = 6378.1 % km. Earth's radius
q = r+35786 % km. distance of geosynchronous orbit from center of earth
e = 23+(56/60) % hours. Period of geosynchronous orbit.
z = q*((14+(40/60))/e)^(2/3) % Distance from satellite to center of earth
answer = z-r
\end{verbatim}

\item A satellite orbits the earth in a circular orbit. It passes directly over the
North and South poles and completes an orbit every 14 hours 40 minutes.
On one orbit going southward it was directly above the earth location
$40^{\circ}$ N,
$10^{\circ}$ W at 1:00 PM EST. At what time will it next cross the
plane that contains
the circle of latitude $40^{\circ}$ N, and what will be its longitude?
(An answer will be marked correct if the time is within 5 minutes of the
correct answer, and the longitude is within 3 degrees.)

{\bf Answer:} Time: 2:19 PM. Longitude: 29.77 degrees W (= $20^{\circ} \circ
40^{\prime}$ W)

{\bf Matlab:}
\begin{verbatim}
% Let c be the center of the earth. 
% Let P be the plane containing the circle of 40 degrees latitude.
% Let s be the position of the satellite when it crosses P 
% Let a be the point where the line from c to the North Pole crosses P.
% Let r be the radius of the earth and let h be the height of the satellite 

% cas is a right triangle. |ca| = r*sin(40 degrees). |cs| = r+h
% So the angle acs = arcsin(|ca|/|cs|). 
% The angle traversed since 1 PM is 40-(90-acs)
% The rest of the calculation is as in problem 29

r = 6378.1 % km. Earth's radius
h = 24042 % height of satellite over Earth's surface. Problem 30
w = 14+(40/60) % period of satellite
d = 23+(56/60) % sidereal day
deg2rad = pi/180
ca = r*sin(40*deg2rad)
cs = r+24042 % Problem 30
acs = asin(ca/cs)
at = 40-(acs/deg2rad) % angle traversed in degrees
lat = 10 + (w/d)*at % latitude
t = 1 + w*(at/360) % time
\end{verbatim}

\end{enumerate}

\section*{Appendix B: The ``Calculation-Free'' test set}
\label{secCalculationFree}

\begin{enumerate}
\subsection*{Eclipse problems}

\item An astronaut is standing in the Sea of Tranquility during what on 
earth is called a total lunar eclipse. They are looking in the
direction of the earth. What they see is:
\begin{itemize}
\item[A.] The surface of the moon, illuminated by earth light.
\item[B.] The night side of the earth, occluding the sun.
\item[C.] The surface of the moon, illuminated only by starlight.
\item[D.] The surface of the moon, illuminated by the sun.
\item[E.] The sun.
\item[F.] The day side of the earth, with a small circular shadow moving 
quickly over it.
\item[G.] The night side of the earth. The sun is somewhere else entirely.
\item[H.] A starry sky. Neither the sun, the earth, or the surface of the moon
is in the field of view.
\end{itemize}
{\bf Answer:}
B. The night side of the earth, occluding the sun.

\item An astronaut is standing in the Sea of Tranquility during what on 
earth is called a total lunar eclipse. They are looking in the
direction of the sun. What they see is:
\begin{itemize}
\item[A.] The surface of the moon, illuminated by earth light.
\item[B.] The night side of the earth, occluding the sun.
\item[C.] The surface of the moon, illuminated only by starlight.
\item[D.] The surface of the moon, illuminated by the sun.
\item[E.] The sun.
\item[F.] The day side of the earth, with a small circular shadow moving 
quickly over it.
\item[G.] The night side of the earth. The sun is somewhere else entirely.
\item[H.] A starry sky. Neither the sun, the earth, or the surface of the moon
is in the field of view.
\end{itemize}
{\bf Answer:}
B. The night side of the earth, occluding the sun.

\item An astronaut is standing in the Sea of Tranquility during what on 
earth is called a total solar eclipse. They are looking in the
direction of the earth. What they see is:
\begin{itemize}
\item[A.] The surface of the moon, illuminated by earth light.
\item[B.] The night side of the earth, occluding the sun.
\item[C.] The surface of the moon, illuminated only by starlight.
\item[D.] The surface of the moon, illuminated by the sun.
\item[E.] The sun.
\item[F.] The day side of the earth, with a small circular shadow moving 
quickly over it.
\item[G.] The night side of the earth. The sun is somewhere else entirely.
\item[H.] A starry sky. Neither the sun, the earth, or the surface of the moon
is in the field of view.
\end{itemize}
{\bf Answer:}
F. The day side of the earth, with a small circular shadow moving 
quickly over it.

\item An astronaut is standing in the Sea of Tranquility during what on 
earth is called a total solar eclipse. They are looking in the
direction of the sun. What they see is:
\begin{itemize}
\item[A.] The surface of the moon, illuminated by earth light.
\item[B.] The night side of the earth, occluding the sun.
\item[C.] The surface of the moon, illuminated only by starlight.
\item[D.] The surface of the moon, illuminated by the sun.
\item[E.] The sun.
\item[F.] The day side of the earth, with a small circular shadow moving 
quickly over it.
\item[G.] The night side of the earth. The sun is somewhere else entirely.
\item[H.] A starry sky. Neither the sun, the earth, or the surface of the moon
is in the field of view.
\end{itemize}
{\bf Answer:}
A. The surface of the moon, illuminated by earth light.

\item An astronaut is standing on the so-called ``dark side of the moon'', 
during what on 
earth is called a total lunar eclipse. They are looking in the
direction of the earth. What they see is:
\begin{itemize}
\item[A.] The surface of the moon, illuminated by earth light.
\item[B.] The night side of the earth, occluding the sun.
\item[C.] The surface of the moon, illuminated only by starlight.
\item[D.] The surface of the moon, illuminated by the sun.
\item[E.] The sun.
\item[F.] The day side of the earth, with a small circular shadow moving 
quickly over it.
\item[G.] The night side of the earth. The sun is somewhere else entirely.
\item[H.] A starry sky. Neither the sun, the earth, or the surface of the moon
is in the field of view.
\end{itemize}
{\bf Answer:}
C. The surface of the moon, illuminated only by starlight.

\item An astronaut is standing on the so-called ``dark side of the moon'', 
during what on 
earth is called a total lunar eclipse. They are looking in the
direction of the sun. What they see is:
\begin{itemize}
\item[A.] The surface of the moon, illuminated by earth light.
\item[B.] The night side of the earth, occluding the sun.
\item[C.] The surface of the moon, illuminated only by starlight.
\item[D.] The surface of the moon, illuminated by the sun.
\item[E.] The sun.
\item[F.] The day side of the earth, with a small circular shadow moving 
quickly over it.
\item[G.] The night side of the earth. The sun is somewhere else entirely.
\item[H.] A starry sky. Neither the sun, the earth, or the surface of the moon
is in the field of view.
\end{itemize}
{\bf Answer:}
C. The surface of the moon, illuminated only by starlight.

\item An astronaut is standing on the so-called ``dark side of the moon'', 
during what on 
earth is called a total solar eclipse. They are looking in the
direction of the earth. What they see is:
\begin{itemize}
\item[A.] The surface of the moon, illuminated by earth light.
\item[B.] The night side of the earth, occluding the sun.
\item[C.] The surface of the moon, illuminated only by starlight.
\item[D.] The surface of the moon, illuminated by the sun.
\item[E.] The sun.
\item[F.] The day side of the earth, with a small circular shadow moving 
quickly over it.
\item[G.] The night side of the earth. The sun is somewhere else entirely.
\item[H.] A starry sky. Neither the sun, the earth, or the surface of the moon
is in the field of view.
\end{itemize}
{\bf Answer:}
D. The surface of the moon, illuminated by the sun.

\item An astronaut is standing on the so-called ``dark side of the moon'' 
during what on 
earth is called a total solar eclipse. They are looking in the
direction of the sun. What they see is:
\begin{itemize}
\item[A.] The surface of the moon, illuminated by earth light.
\item[B.] The night side of the earth, occluding the sun.
\item[C.] The surface of the moon, illuminated only by starlight.
\item[D.] The surface of the moon, illuminated by the sun.
\item[E.] The sun.
\item[F.] The day side of the earth, with a small circular shadow moving 
quickly over it.
\item[G.] The night side of the earth. The sun is somewhere else entirely.
\item[H.] A starry sky. Neither the sun, the earth, or the surface of the moon
is in the field of view.
\end{itemize}
{\bf Answer:}
E. The sun.

\subsection*{Distance combination problems}
\item Joe says that he lives 10 miles from the Atlantic, that Beth lives 
10 miles from the Atlantic, and that he and Beth live 3000 miles apart.
Is it possible that Joe is telling the truth? {\bf Answer} Yes 

\item Joe says that he lives 10 miles from New York City, that Beth lives 
10 miles from New York City, and that he and Beth live 3000 miles apart.
Is it possible that Joe is telling the truth? {\bf Answer} No.

\item Joe says that he lives 10 miles from Lake Michigan, that Beth lives 
10 miles from Lake Michigan, and that he and Beth live 100 miles apart.
Is it possible that Joe is telling the truth? {\bf Answer} Yes

\item Joe says that he lives 10 miles from Walden Pond, that Beth lives 
10 miles from Walden Pond, and that he and Beth live 100 miles apart.
Is it possible that Joe is telling the truth? {\bf Answer} No,

\item Joe says that he lives 100 miles from Walden Pond, that Beth lives 
100 miles from Lake Michigan, and that he and Beth live 10 miles apart.
Is it possible that Joe is telling the truth? {\bf Answer} No.

\item Joe says that he lives 1000 miles from Walden Pond, that Beth lives 
1000 miles from Lake Michigan, and that he and Beth live 10 miles apart.
Is it possible that Joe is telling the truth? {\bf Answer} Yes

\item Joe says that he lives 10 miles from Lake Huron, that Beth lives 
10 miles from Lake Michigan, and that he and Beth live 10 miles apart.
Is it possible that Joe is telling the truth? {\bf Answer} Yes.

\item Is there a point in the Nile River that is exactly 827 miles from the
Al-Azhar Mosque in Cairo? {\bf Answer} Yes.

\item Is there a point in the Nile River that is exactly 8562 miles from the
Al-Azhar Mosque in Cairo? {\bf Answer} No.

\item Is there a point x in the Danube River and a point y in the Rhine River
that are exactly 12 miles apart? {\bf Answer} No.

\item Is there a point x in the Danube River and a point y in the Rhine River
that are exactly 429 miles apart? {\bf Answer} Yes.

\item Is there a point x in the Danube River and a point y in the Rhine River
that are exactly 8738 miles apart? {\bf Answer} No.

\subsection*{Clockwise vs. counterclockwise}
If you have a map that shows Chicago, New York City, and Atlanta, and 
you draw a circle through the three of them,
then the sequence $\la$Chicago; New York; Atlanta$\ra$ is in clockwise order. 
The sequence
$\la$New York; Chicago; Atlanta$\ra$ on the other hand, is in 
counterclockwise order.

For each of the following, state whether it is in clockwise or counterclockwise
order:

\item Caracas, Venezuela; \hspace{1em}  Amarillo, Texas; 
\hspace{1em} Quebec, Quebec. \\ {\bf Answer:} Clockwise.
\item New Orleans, Louisiana;  \hspace{1em} Springfield, Illinois; 
\hspace{1em} Jacksonville, Florida. \\ {\bf Answer:} Clockwise.
\item Kingston, Jamaica; \hspace{1em}  Fresno, California; 
\hspace{1em} Albany, New York. \\ {\bf Answer:} Clockwise.
\item Indianapolis, Indiana; \hspace{1em}  Wichita, Kansas; 
\hspace{1em} Montgomery, Alabama. \\ {\bf Answer:} Counterclockwise.
\item Denver, Colorado; \hspace{1em}  Houston, Texas; 
\hspace{1em} Indianapolis, Indiana. \\ {\bf Answer:} Counterclockwise.
\item Seattle, Washington; \hspace{1em}  New York, New York; 
\hspace{1em} Mazatlan, Mexico. \\ {\bf Answer:} Clockwise.
\item Spokane, Washington; \hspace{1em}  Mexico City, Mexico; 
\hspace{1em} Virginia Beach, Virginia. \\ {\bf Answer:} Counterclockwise.
\item Sitka, Alaska;  \hspace{1em} Oakland, California; 
\hspace{1em} Albany, New York. \\ {\bf Answer:} Counterclockwise.
\item Pierre, South Dakota; \hspace{1em}  Eastport, Maine; 
\hspace{1em} Columbia, South Carolina. \\ {\bf Answer:} Clockwise.
\item Toronto, Ontario; \hspace{1em}  El Paso, Texas; 
\hspace{1em} Key West, Florida. \\ {\bf Answer:} Counterclockwise.
\item Sitka, Alaska; \hspace{1em}  Omaha, Nebraska; 
\hspace{1em} Long Beach, California. \\ {\bf Answer:} Clockwise.
\item Edmonton, Alberta; \hspace{1em} New Orleans, Louisiana;
\hspace{1em} Washington DC. \\ {\bf Answer:} Counterclockwise

\subsection*{Sorting problems}
Sorting problems should be scored in terms of the number of pairs in order
divided by the total number of pairs.  The order of answers has been randomized.
\item Sort the items below in increasing order of mass:
\begin{itemize}
\item[A.] Pablo Picasso, when five years old.
\item[B.] The Great Sphinx
\item[C.] An atom of uranium.
\item[D.] The planet Mercury.
\item[E.] A Toyota Corolla
\item[F.] The polio virus.
\item[G.] A hamster.
\item[H.] Emily Dickinson, when twenty-three years old.

{\bf Answer:} C, F, G, A, H, E, B, D.
\end{itemize}

\item Sort the events below by duration:
\begin{itemize}
\item[A.] The lifetime of Marie Antoinette.
\item[B.] The Precambrian period.
\item[C.] The first performance of Beethoven's seventh symphony.
\item[D.] Lincoln speaking the Gettysburg address.
\item[E.] The Hundred Years' War.
\item[F.] The reign of Queen Victoria.
\item[G.] The existence of the species of passenger pigeons (ending with
the death of ``Martha'').
\item[H.] The battle of Gettysburg.
\item[I.] The existence of legal slavery in what is now the United States.
\item[J.] The reign of Marie Antoinette.
\item[K.] The lifetime of Joan of Arc.

{\bf Answer:} D, C, H, K, J, A, F, E, I, G, B 
\end{itemize}

(Note: The durations of K, the lifetime of Joan of Arc, and J, the reign of
Marie Antoinette, are very close -- about 19 years. Moreover, the starting
date of the former is not known precisely, and the ending date of the latter
is debatable.  Either order
should be accepted as correct for these. This was included by misdesign.)

\item Sort the events below by starting date:
\begin{itemize}
\item[A.] The lifetime of Marie Antoinette.
\item[B.] The Precambrian period.
\item[C.] The first performance of Beethoven's seventh symphony.
\item[D.] Lincoln speaking the Gettysburg address.
\item[E.] The Hundred Years' War.
\item[F.] The reign of Queen Victoria.
\item[G.] The existence of the species of passenger pigeons (ending with
the death of ``Martha'').
\item[H.] The battle of Gettysburg.
\item[I.] The existence of legal slavery in what is now the United States.
\item[J.] The reign of Marie Antoinette.
\item[K.] The lifetime of Joan of Arc.
\end{itemize}

{\bf Answer:} B, G, E, K, I, A, J, C, F, H, D  

\item Sort the events below by ending date:
\begin{itemize}
\item[A.] The lifetime of Marie Antoinette.
\item[B.] The Precambrian period.
\item[C.] The first performance of Beethoven's seventh symphony.
\item[D.] Lincoln speaking the Gettysburg address.
\item[E.] The Hundred Years' War.
\item[F.] The reign of Queen Victoria.
\item[G.] The existence of the species of passenger pigeons (ending with
the death of ``Martha'').
\item[H.] The battle of Gettysburg.
\item[I.] The existence of legal slavery in what is now the United States.
\item[J.] The reign of Marie Antoinette.
\item[K.] The lifetime of Joan of Arc.
\end{itemize}

{\bf Answer:} B, K, E, J, A, C, H, D, I, F, G.

\item Sort these entities in increasing order of distance from the painting
``The Mona Lisa''.
\begin{itemize}
\item[A.] Omaha Beach
\item[B.] Phobos, the moon of Mars.
\item[C.] The Dome of the Rock
\item[D.] The Washington Monument
\item[E.] The Sistine Chapel
\item[F.] Leonardo da Vinci's painting, ``St. John the Baptist''.
\item[G.] The Tomb of Napoleon.
\item[H.] The Rosetta Stone
\item[I.] The Parthenon
\item[J.] Mount Fujiyama
\item[K.] The Great Barrier Reef.
\end{itemize}
{\bf Answer:} F, G, A, H, E, I, C, D, J, K, B

\item Sort these people by increasing order of the distance of their
birthplace from the birthplace of Frederic Chopin.
\begin{itemize}
\item[A.] Gandhi
\item[B.] Mao
\item[C.] Cleopatra
\item[D.] Kafka
\item[E.] Mohammed
\item[F.] Cate Blanchett
\item[G.] Napoleon
\item[H.] Marie Curie
\end{itemize}

{\bf Answer:} H, D, G, E, A, B, F.

\item Sort these people by increasing order of the distance of their birthplace
from the birthplace of Abraham Lincoln.
\begin{itemize}
\item[A.] Ruth Bader Ginsburg
\item[B.] The Duke of Wellington
\item[C.] Ulysses Grant
\item[D.] Vladimir Lenin
\item[E.] Jacindra Ardren.
\item[F.] Ho Chi Minh
\item[G.] Kamala Harris
\item[H.] Barack Obama
\item[I.] George Washington
\end{itemize}

{\bf Answer:} C, I, A, G, B, H, D, E, F

Note: It is not at all obvious, without doing the calculation, that Abraham
Lincoln's birthplace (Hodgenville, Ky.) is closer to the Duke of Wellington's
(Dublin) or that it is closer to Jacindra Ardern's birthplace (Hamilton,
New Zealand) than to Ho Chi Minh's (Nghe An province, Vietnam). These 
comparisons were included due to an error in design.


\subsection*{Satellites}
In the following assume that the satellite is moving in a closed
orbit around the Earth
and that the only influence on the satellite's motion  is the
Earth's gravity. Ignore the revolution of the Earth around the sun,
but do not ignore the rotation of the Earth around its axis. If there are two
or more satellites, assume that it is not possible for them to occupy the same
point in space simultaneously.

\item Is it possible to have a satellite such that the point on the earth
underneath the satellite always has longitude $40^{\circ}$ W? \\
{\bf Answer:} Yes (a satellite in geostationary orbit 
that is always over the earth
point on the equator at $30^{\circ}$ S, $40^{\circ}$ W).

\item Is it possible to have a satellite such that the earth point underneath
the satellite always has latitude $40^{\circ}$ S? \\
{\bf Answer:}  No.

\item Is it possible to have a satellite such that the northernmost
earth point underneath the satellite has latitude $40^{\circ}$ N and the
southernmost has latitude $30^{\circ}$ S? \\  
{\bf Answer:}  No.

\item Assume that a satellite passes in its orbit over the North Pole. Which of
the following is true:
\begin{itemize}
\item[A.] It must pass over the South Pole.
\item[B.] It cannot pass over the South Pole.
\item[C.] It might or might not pass over the South Pole.
\end{itemize}
{\bf Answer:} A

\item Suppose that the orbit of a satellite, which is non-circular,
takes it over the North Pole, and
moreover, it is furthest from the earth when it is over the North Pole.
When will it be nearest to the earth?
\begin{itemize}
\item[A.] When it is over the equator.
\item[B.] When it is over the South Pole.
\item[C.] At some point after it has crossed the equator but before it has
passed over the South Pole.
\item[D.] None of the above are necessarily true.
\end{itemize}
{\bf Answer:} B

\item Can two earth satellites have their orbits in planes that are not
equal but are parallel? \\
{\bf Answer:}  No

\item Can two earth satellites have their orbits in planes that are
orthogonal? \\
{\bf Answer:} Yes

\item  Let C be the center of the earth.
Can there be two earth satellites X and Y
such that X, C, and Y always lie in a straight line, with C between X and Y? 
{\bf Answer:} Yes

\item  Let C be the center of the earth.
Can there be two earth satellites X and Y
such that C, X, and Y always lie in a straight line, with X between C and Y?
{\bf Answer:} No.

\item  Let C be the center of the earth.
Can there be two earth satellites X and Y
such that the angle XCY is always a right angle?
{\bf Answer:} Yes.

\item  Let C be the center of the earth.
Can there be two earth satellites X and Y
such that the angle CXY is always a right angle?
{\bf Answer:} No.

\item  Let C be the center of the earth.
Can there be two earth satellites X and Y
such that the angle CXY is always 60 degrees?
{\bf Answer:} Yes.

\item  Let C be the center of the earth.
Can there be three earth satellites X, Y, and Z such that C, X, Y, and Z
are always coplanar? \\
{\bf Answer:} Yes.

\item  Let C be the center of the earth.
Can there be three earth satellites X, Y, and Z
such that the lines CX, CY, and CZ are always all pairwise orthogonal? \\
{\bf Answer:} No.
\end{enumerate}

\section*{Appendix C: The ``Motivated Numerical'' test set}
\label{secMotivated}

\begin{enumerate}
\item 
{\bf Question:}
If the earth collapsed to a black hole, how big would the black hole
be? Please include all calculations.

{\bf Answer:} 8.87 millimeters.

\item
If you fell into the black hole at the center of the Milky Way, how
long would you have before hitting the singularity? Please include all
calculations.

{\bf Answer:} 66.7 seconds.

\item Consider the first binary black hole system discovered by LIGO.
Roughly how close to that system would a person have to have been,
before they were killed by the gravitational waves? Please include all
calculations.

{\bf Answer:} So close that you would be killed by the tidal forces rather than
by the gravitational waves.

\item
How far off from 10! is Stirling's approximation? Please include all calculations.

{\bf Answer:} 30,104.

\item
How high would an airplane have to be, before you could notice 10
degrees of the earth's curvature when looking out the window?
Please include all calculations.

{\bf Answer:} 98.4 km.

{\bf Matlab}
\begin{verbatim}
r = 6378 % Radius of earth in kilometers
d = r/cos(10*pi/180)  % distance from center of earth to airplane
answer = d-r %height above earth's surface
\end{verbatim}

\item
Approximately how much time would a commercial airliner save in going from New York 
to Tel Aviv, if it could go in a straight line, through
a tunnel in the earth, at the same speed as usual? Please include all
calculations.

{\bf Answer:} About 52 minutes, assuming a speed of 550 mph.

{\bf Matlab:}
\begin{verbatim}
nyclat = (40+42/60+46/3600)*pi/180
nyclong = -(74+22/3600)*pi/180
talat = (32+8/60)*(pi/180)
talong = (34+78/60)*(pi/180)
earthRadius = 3959 % miles
nycDirection = [cos(nyclat)*sin(nyclong), cos(nyclat)*cos(nyclong), sin(nyclat)]  % unit vector
taDirection = [cos(talat)*sin(talong), cos(talat)*cos(talong), sin(talat)]
tunnel = earthRadius*norm(taDirection-nycDirection)
flight = earthRadius*acos(dot(nycDirection,taDirection))
timeSaved = (flight-tunnel)/550 % assuming 550 mph.
\end{verbatim}

\item
 For what fraction of the lifetime of the universe has there been
life in it?  Give upper and lower bounds.
Please include all calculations.

{\bf Answer:} Between 25\% and 99.93\%.

{\bf Explanation:} The age of the universe is taken to be 13.8 billion years.
It is generally accepted that life emerged on earth at least 3.5 billion years
ago, and it is speculated that it might have been possible for life to appear
in the universe about 10 years after the Big Bang.

\item
Approximately how long would it take to transmit an entire human
genome over a standard WiFi connection?
Please include all calculations.

{\bf Answer:} 4 minutes.

{\bf Explanation:} The human genome has about 3 billion base pairs = 6 billion
bits. Assuming a 25Mbs Wifi connection, that gives 240 seconds = 4 minutes.

\item
Approximately how large would an asteroid have to be, in diameter
(assume it's approximately spherical), before an Olympic high jumper
could no longer reach escape velocity by jumping off it?
Please include all calculations.

{\bf Answer} About 10 km.

{\bf Explanation:} The kinetic energy needed for escape velocity from the
surface of a planet is inversely proportional to the radius and proportional
to the mass of the planet, which in turn is proportional to the
density and radius cubed.
So over planets of constant density energy is proportional to the radius
squared and density.
So the velocity is proportional to the radius and square root of the density.
So the radius as a proportional to the velocity and inversely proportional
to the square root of the density.

The world record for high jump is 2.45 meters, set by Javier Sotomayor in 1993.
The jumper's center of mass starts about half way up his body, around .9 m.
(Sotomayor is 193 cm tall.) At the apex of the jump, it is perhaps .3m above
the bar. So the jump raises the center of mass by about 1.8 meters. The initial
velocity is therefore $v = \sqrt{2gh} \approx 6$ m/s.

Escape velocity from earth is 11 km/s and the diameter of earth is 12,750 km.
The densities of asteroids is not at all well known (see
\href{https://en.wikipedia.org/wiki/Standard_asteroid_physical_characteristics#Density}{the Wikipedia article, ``Standard asteroid physical characteristics''}
but a reasonable estimate might be about 4/9 the density of earth.

So the diameter of the asteroid would be about
$12750 \cdot (6/11000) \cdot (3/2) \
\approx 10$ km.

\item
What is the length of the $y=x^2$ parabola in the region $-1 \leq x \leq 1$?
Please include all calculations.

{\bf Answer:} 
\[ \int_{-1}^{1} \sqrt{1+4x^{2}} \: dx = \sqrt{5} + \frac{1}{2}\sinh^{-1}(2) =
2.95789 \]

\item
Approximately how large a supply of antimatter would be needed, in
order to propel a spacecraft with the mass of the International Space
Station into orbit around Proxima Centauri, in one year as experienced
by its crew?
Please include all calculations.

{\bf Answer:} 706,000 kg

{\bf Explanation:} \\
We'll assume that the Space Station is accelerated rapidly to its cruising
speed, and we'll ignore the need to decelerate when it reaches Proxima 
Centauri. 

The first step is to compute the velocity. We'll use the speed of light as the 
unit of speed and a year as the unit of time. Proxima Centauri is 4.25 light years
away. If the spaceship travels at speed $v$, then as measured from earth,
the time required $\tau =  4.25/v$. Because of time dilation, this appears to
the crew as $\tau \sqrt{1-v^{2}}$. So we have the equation 
$1=(4.25/v)   \sqrt{1-v^{2}}$, so
$1=(4.25^{2}/v^{2})(1-v^{2})$, so 
$v^{2} - 18.0625(1-v^{2})$, so $v^{2} = 18.0625/19.062 = 0.9475$ and
$v=0.9734$.

If the Space Station has mass $M$, then its total energy at speed $v$ is
$M/\sqrt{1-v^{2}}$, so the additional energy is 
$M/\sqrt{1-v^{2}} - M$ ($c$ is still 1). If we annihilate mass $m$ of
antimatter with mass $m$ of matter, then that liberates $2m$ of energy
We have $2m = M\cdot ((1/\sqrt{1-v^{2}})-1) = 3.3661M$ so $m=1.683M$. Taking $m$
to be 420,000 kg, that gives a value of 706,000 kg of antimatter.

\item
Approximately how many errors will a standard laptop suffer over
its lifetime, due to cosmic rays hitting the microchip?
Please include all calculations.

{\bf Answer:} Estimates vary widely, but one commonly cited figure is
about 1 error per 256 megabytes of RAM per month. Assuming an 8 GByte laptop and
a 5 year lifespan, that will total 1920 errors.

\item
What is the approximate probability that a randomly-chosen
100-digit integer is prime?
Please include all calculations.

{\bf Answer:} Approximately $1/\log_{e}(5.5 \cdot 10^{99}) =
1/(\log_{e}(5.5) + 99 \cdot \log_{e}(10)) = 0.00435$. That should be accurate to within 1\%.

\item
What is the probability that a randomly-chosen 100 $\times$ 100 matrix,
over the finite field $F_{2}$, is invertible?
Please include all calculations.

{\bf Answer:} 0.289

{\bf Explanation:}  Calculate the probability that
the 2nd row is linearly independent from the first, then the
probability that the 3rd row is outside the span of the first two, the
probability that the 4th row is outside the span of the first three,
etc. and multiplying them all together. In backward order this is

$1/2 \cdot (3/4)  \cdot (7/8) \cdot (15/16) \cdot ... \approx 0.289.$


\item
How does the total weight of all the uranium that humans mined,
compare to the total weight of all the gold that they've mined?
Please include all calculations.

{\bf Answer:} It is estimated that humans have mined about 209,000 tones of
gold and about 2.8 million tons of uranium, a difference of a factor of
about 13.

{\bf Explanation:} The figure for gold comes from 
\href{https://www.gold.org/goldhub/data/how-much-gold}{the World Gold
Council.}  The figure for uranium was estimated by the authors based on the
information on
\href{https://www.world-nuclear.org/information-library/nuclear-fuel-cycle/mining-of-uranium/world-uranium-mining-production.aspx}{this web page of the World
Nucler Association}

\item
What is the Shannon entropy of a positive integer n that's chosen
with probability Pr[n] = $6/(\pi^2 \cdot n^2)$?
Please include all calculations.

{\bf Answer:} 2.362

{\bf Matlab:}
\begin{verbatim}
ent=0;
for i=1:1000000
  p = 6/((pi^2)*(i^2));
  ent = ent - p*log(p);
  sum = sum+p;
end
ent/log(2)
\end{verbatim}

{\bf Revised version of problem:}
On consideration, after running the above, we thought that perhaps the wording
of the problem was unfair to GPT. After all, the answer probably has no
elegant closed-form expression, and it is not easy to get a highly precise
numerical answer (the convergence rate of the na\"{i}ve summation is 
$O(\log(n)/n)$). So we tried the two systems with the following revised 
question:

Compute the Shannon entropy of a positive integer n that's chosen
with probability Pr[n] = $6/(\pi^2 \cdot n^2)$ to 3 digit accuracy. 
Please include all calculations.

\item
Assume that IQs are normally distributed, with a mean of 100 and a
standard deviation of 15.  For which n does the number of people with
IQ n exceed the number of people with IQ n+1 by the maximum amount?
Please include all calculations.

{\bf Answer:} Approximately $n=14.5$ 

{\bf Explanation:}
Let $f(x)$ be the probability density function for the normal distribution
with mean 100.

$f(x) = \frac{1}{\sqrt{2 \pi} \cdot 15} \mbox{exp}(-(x-100)^{2}/2 \cdot 
15^{2})$
At the infinitesimal scale, the change between people with IQ $x$ and people
with IQ $x+\epsilon$ is maximized when the derivative of the probability 
distribution is negative and its magnitude is maximal; thus, at a value $x$
greater than the mean where the second derivative is 0.

The first derivative of $f$,
$f^{\prime}(x) = -((x-100)/15^{2}) f(x)$.

Setting the second derivative of $f$ to be zero, we have,
$f^{\prime \prime} = ((x-100)^{2}/15^{2})^{2}-(1/15^{2})f(x)$, so
$(x-100)^{2} = 15^{2}$, which has two roots: $x=85$, where the first
derivative reaches a maximum positive value and $x=105$ where it reaches
a maximum magnitude negative value.

Now, that does not directly give us the value of $x$ that maximizes 
$f(x)-f(x+1)$, and finding the true maximum would involve solving
an inelegant transcendental equation. 
However, since the third derivative of $f$ is fairly small and
the second derivative is 0, it is safe to assume that picking $x$ and $x+1$
evenly across $x=15$ gives values close to the maximum value. Using binary
search, one can determine that the true maximum is 114.5028.

\item
Suppose a randomized algorithm outputs the correct answer (yes or
no) with probability 0.9.  If we take the majority answer out of 100
independent runs, with what probability will the answer be wrong?
Please include all calculations.

{\bf Answer:} $3.232 \cdot 10^{-24}$.

{\bf Explanation:} Assume that if the results are exactly 50/50, then
you flip a fair coin. Then the exact value is

\[ (1/2) \cdot C(100.50) \cdot 0.9^{50} \cdot 0.1^{50} +
\sum_{i=1}^{50} C(100,50+i) \cdot 0.9^{50-i} \cdot 0.1^{50+i} \]

However, since this is (approximately) 
a geometrically decreasing series with ratio of 1/9,
it suffices to take the first five terms to get three digit accuracy. 

{\bf Matlab:} (Matlab warns that its value may have a relative error of
$10^{-14}$).

\begin{verbatim}
p=nchoosek(100,50)*0.9^50*0.1^50
sum=p/2
for i=1:5
   p=nchoosek(100,50+i)*0.9^(50-i)*0.1^(50+i)
   sum = sum+p
end
sum 
\end{verbatim}

\item
In a Manhattan-like two-dimensional grid, how many distinct
shortest paths are there that go from a fixed starting point to a
fixed endpoint 5 blocks to the north and 5 blocks to the east?
Please include all calculations.

{\bf Answer:} 252.

{\bf Explanation:} You have to take 10 steps: 5 of them north and 5 of 
them east. They can be in any order. So the number is C(10,5) = 252.

\item
When 3-dimensional spheres are packed in their densest possible
configuration, what percentage of the space is empty?
Please include all calculations.

{\bf Answer:} $1-\pi/3\sqrt{2} = 0.2595$. See the
Wikipedia article 
``\href{https://en.wikipedia.org/wiki/Kepler_conjecture}
{Kepler conjecture}''.

\end{enumerate}
\end{document}